\documentclass[journal]{IEEEtran}
\usepackage{array}
\usepackage{cite}
\usepackage{caption}
\usepackage{amsmath,graphicx}
\usepackage{epstopdf}
\usepackage{xcolor}
\usepackage{amssymb}
\usepackage{pifont}
\usepackage{booktabs}
\usepackage{array}
\usepackage{tabu}
\usepackage{booktabs}
\usepackage{graphicx}
\usepackage{subfigure}
\usepackage{bm}
\usepackage{multirow}
\usepackage{graphicx}
\usepackage{color}
\usepackage{float}
\usepackage{stfloats}
\usepackage[misc]{ifsym}
\usepackage{stmaryrd}
\usepackage{grffile}
\usepackage{makecell}
\usepackage{threeparttable}
\usepackage{bm}
\usepackage{enumitem}
\usepackage{enumerate}
\usepackage[pagebackref=true,breaklinks=true,letterpaper=true,colorlinks,bookmarks=false]{hyperref}

\begin{document}
	
	\captionsetup[figure]{labelformat={default},labelsep=period,name={Fig.}}
	
	\title{RRCANet: Recurrent Reusable-Convolution Attention Network for Infrared Small Target Detection}
	
	\author{{Yongxian Liu, Boyang Li, Ting Liu, Zaiping Lin, Wei An}
    
	   \thanks{The work was supported by the National Natural Science Foundation of China under Grant 62501618, and Grant 62505266. (\textit{Corresponding author: Zaiping Lin; Wei An.})}	
		\thanks{Yongxian Liu, Boyang Li, Zaiping Lin and Wei An are with the College of Electronic Science and Technology, National University of Defense Technology (NUDT), Changsha 410073, China. (e-mail: yongxian23@nudt.edu.cn; liboyang20@nudt.edu.cn, linzaiping@nudt.edu.cn, anwei@nudt.edu.cn.).
		
		Ting Liu is with the College of Automation and Electronic Information, Xiangtan University (XTU), Xiangtan 411105, China. (e-mail: liuting@nudt.edu.cn)
		
		Our code will be available at \url{https://github.com/yongxianLiu/RRCANet} soon
		}
	}

	\maketitle
	
	\begin{abstract}
		Infrared small target detection is a challenging task due to its unique characteristics (e.g., small, dim, shapeless and changeable). Recently published CNN-based methods have achieved promising performance with heavy feature extraction and fusion modules. To achieve efficient and effective detection, we propose a recurrent reusable-convolution attention network (RRCA-Net) for infrared small target detection. Specifically, RRCA-Net incorporates reusable-convolution block (RuCB) in a recurrent manner without introducing extra parameters. With the help of the repetitive iteration in RuCB, the high-level information of small targets in the deep layers can be well maintained and further refined. Then, a dual interactive attention aggregation module (DIAAM) is proposed to promote the mutual enhancement and fusion of refined information. In this way, RRCA-Net can both achieve high-level feature refinement and enhance the correlation of contextual information between adjacent layers. Moreover, to achieve steady convergence, we design a target characteristic inspired loss function (DpT-k loss) by integrating physical and mathematical constraints. Experimental results on three benchmark datasets (e.g. NUAA-SIRST, IRSTD-1k, DenseSIRST) demonstrate that our RRCA-Net can achieve comparable performance to the state-of-the-art methods while maintaining a small number of parameters, and act as a plug and play module to introduce consistent performance improvement for several popular IRSTD methods.
		
	\end{abstract}
	
	
	\section{Introduction}\label{introduction}
	\IEEEPARstart{I}{nfrared} search and tracking system (IRSTS) has the advantage of high detection accuracy and all-weather visibility. It has lots of important applications in many fields such as fire alarm\cite{1-fire-alarm}, maritime surveillance \cite{2-Maritime-Surveillance}, medical diagnosis \cite{3-medical-diagnosis} and precise guidance \cite{4-anti-miss}. Compared to generic object, infrared targets have small size, low signal-to-clutter ratio (SCR) and sparse intrinsic feature. It is challenging to separate such small targets from clutter background.
	
	Traditional detection for infrared small targets are model-driven methods and include three main categories: filtering-based approaches \cite{5-tophat,6-maxmedian}, local-contrast-based approaches \cite{7-WSLCM,8-TLLCM}, low-rank-sparse-decomposition-based approaches \cite{8-IPI,10-NRAM,10-RIPT,9-PSTNN}. However, traditional methods are significantly influenced by expert prior knowledge and handcrafted features, resulting in poor detection performance and generalization ability.
	
	\begin{figure}
		\centering
		\includegraphics[width=8.8cm]{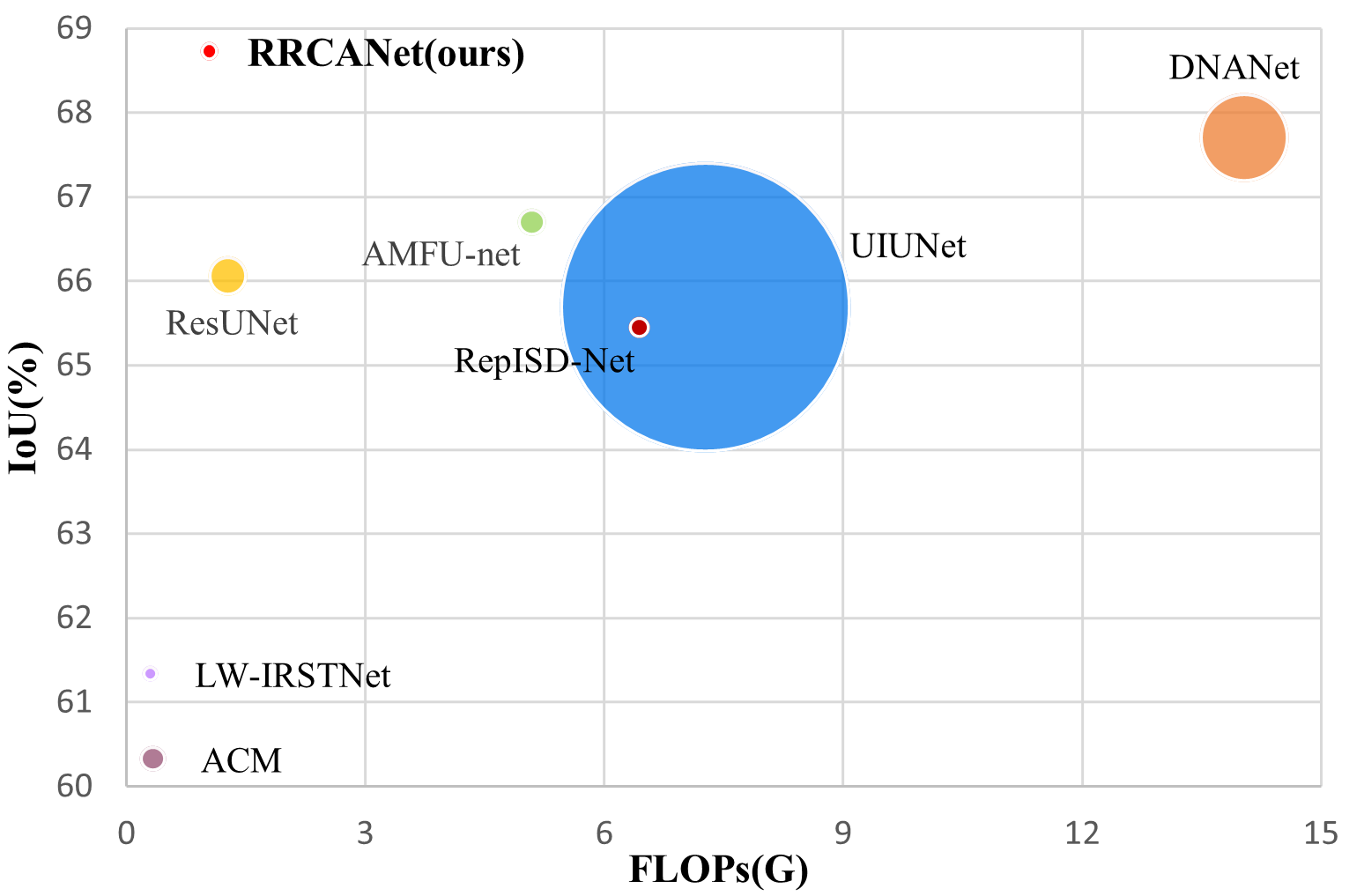}
		\caption{{Comparison between our RRCANet and recently published methods in term of IoU, FLOPs and number of parameters. Larger circle represents more parameters.}}\label{Fig_1}
	\end{figure}

	Deep-learning-based methods can interpret the high-level semantic correlation between small targets and clutter background with a data-driven manner and thus achieve high-precision detection. Dai et al. \cite{11-ACM} proposed an asymmetric contextual module (ACM) to combine top-down global context feedback with a bottom-up modulated path for highlighting small targets. Then, Li et al. \cite{12-DNANet} designed the dense nested interaction module (DNIM) and cascade channel and spatial attention module (CSAM) to preserve small target information and enhance multi-scale features. Wu et al. \cite{13-UIUNet} embedded a tiny U-Net into a larger U-Net backbone to form nested structure, which effectively addresses the problem that feature resolution decreases with the increase of network depth in neural networks. Next, ISNet \cite{13-isnet} extracted edge and shape features to reconstruct targets, effectively improving the detection accuracy. Recently, Yuan et al. \cite{14-sctransnet} proposed a spatial-channel cross transformer block (SCTBs) to enlarge the semantic differences between targets and clutter at all levels. Although these methods have achieved promising performance, they mainly rely on highly stacked network structure, dense connections and heavy multi-scale feature fusion. Different from previous methods with massive modules and heavy computation burden,  we aim to design a more lightweight manner to achieve feature refinement. A recurrent reusable convolution network is proposed to increase the detection performance but maintain comparable parameters with previous method.

	\begin{figure}
		\centering
		\includegraphics[width=8.8cm]{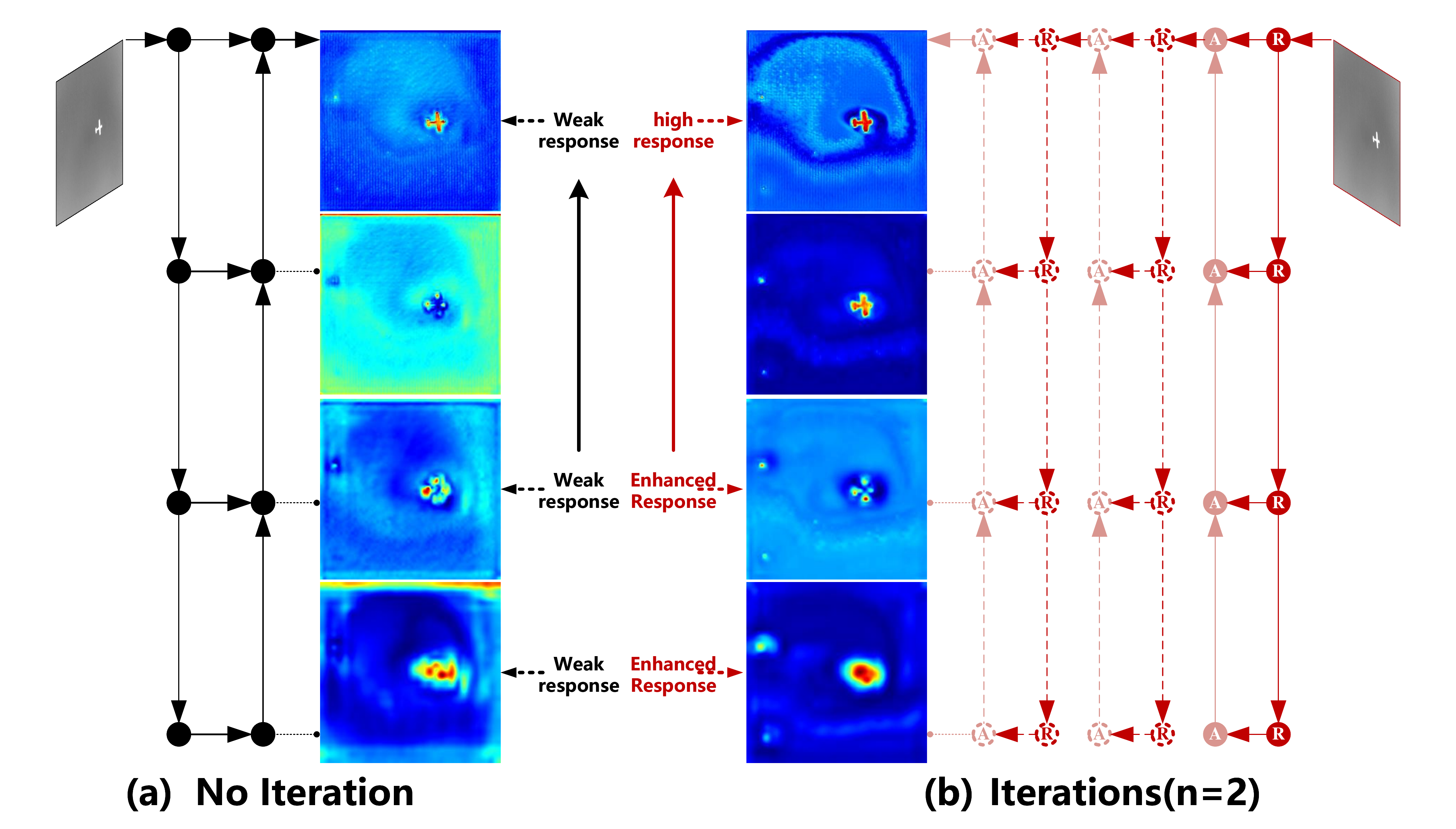}
		\caption{Representation of small targets in CNN with recurrent iteration of (a) RRCA-Net without iteration (b) RRCA-Net with n=2.}\label{Fig_22}
	\end{figure}

	Inspired by reusing mechanism in medical segmentation \cite{27-Bio-Net}, we introduce a lightweight recurrent reusable-convolution attention network (RRCA-Net) to achieve better performance without introducing extra parameters. Specifically, we design a reusable-convolution block (RuCB) to progressively refine semantic features extraction of small targets via iterative learning manner. 
    Visualization result in Fig.~\ref{Fig_22}\textcolor{red}{(b)} is an intuitive example. High-level features obtained by the first encoder-decoder structure are subsequently fed into the second one again for the secondary extraction and aggregation. With the help of this simple reusing manner, our network can achieve feature refinement after repeated convolution operation.  Based on the above RuCB, we  design a dual interactive attention aggregation module (DIAAM) to achieve a attention-based multi-scale feature fusion. Moreover, inspired by the small target characteristics, we integrate physical and mathematical constrains to both maximize the intersection of true target and predicted target, making the network more focus on hard example. In summary, the contributions of our paper can be summarized as follows.
	
	\begin{itemize}
	\item[1)] We propose a simple yet effective RRCA-Net to achieve high-performance detection by reusing convolution operators in a recurrent manner. Without introducing additional parameters, our network can refine and integrate the features extracted by multiple iterations, and act as a plug and play module to introduce consistent performance improvement for several popular IRSTD methods.
	\item[2)] An RuCB and a DIAAM are proposed to achieve high-level feature refinement and enhance the contextual information aggregation capability.
	\item[3)] We develop a small target characteristic inspired DpT-k loss to facilitate network convergence by introducing physical and mathematical constraints. Experimental results on three benchmark datasets (e.g. NUAA-SIRST, IRSTD-1k and DenseSIRST) show that our methods achieve comparable detection performance with fewer  parameters. (see Fig.~\ref{Fig_1}).
	\end{itemize}
	
	The rest of this article is organized as follows. Section \ref{SecRelatedWork} briefly reviews infrared small target detection (IRSTD) methods, lightweight detection networks, and specially designed loss functions. Section \ref{SecMethodology}  introduces the architecture of our RRCA-Net and our DpT-k loss. Experimental results on three single-frame dataset are discussed in Section \ref{SecExperiment}. Finally, Section \ref{SecConclusion} concludes this article.
	
	\section{Related Work}\label{SecRelatedWork}
	
	In this section, we mainly review the main work on IRSTD.
	
	\subsection{Infrared Small Target Detection}
	IRSTD is a challenging task for infrared search and tracking system. In the early years, traditional methods extracted handcrafted features (e.g., contrast and gradient) to detect targets. Typical traditional detection methods, including Top-Hat \cite{5-tophat} and Max-Median \cite{6-maxmedian} adopted sepcially-designed filters to suppress the uniform  background clutter. Then, inspired by the human visual system, local-contrast-based methods \cite{7-WSLCM,8-TLLCM} achieved good performance by calculating the contrast information to distinguish enhanced targets and surrounding background. After that,  low-rank sparse decomposition-based methods \cite{8-IPI,10-NRAM,10-RIPT,9-PSTNN} transformed the small target detection into sparse matrix recovery to increase the performance toward complex background. In summary, these traditional methods heavily rely on prior knowledge and handcrafted features, which is difficult to adapt them to those scenarios with dramatic scene change (e.g., target size, target shape, SCR, and clutter background).
	
	Deep-learning-based methods can learn robust representation via a data-driven manner. Reachers have introduced a variety of neural networks to IRSTD. Many detection methods \cite{11-ACM,12-DNANet,13-UIUNet} were based on CNN and designed many target characteristic-inspired modules to improve detection performance, including spatial attention, dense connection, module nesting and so on \cite{dim2clear, fc3net}. Highly stacked modules and heavy feature fusion strategies significantly improved the detection performance but introduced huge  computation burden. However, CNN-based methods lack the long-range feature aggregation ability. To handle this problem, MTU-Net \cite{32-mtu-Net} applied a multi-level Vision Transformer (ViT)-CNN hybrid encoder to extract and integrate local details and global information. Based on neural ordinary differential equations, RKformer \cite{rkformer} implemented the Runge-Kutta transformer blocks, which can effectively suppress noise. SCTransNet \cite{14-sctransnet} leveraged SCTBs on skip connections to model effectually long-range information. It encoded the semantic difference between targets and background to enhance feature representation. Considering the high computational complexity, Chen et al. \cite{mim} proposed the nested Mamba network (MiM-ISTD) to achieve fast and low-memory detection. Besides, several new paradigms of neural networks have also been introduced into IRSTD. Zhang et al. \cite{irsam} redesigned the segment anything model (SAM) for IRSTD task, demonstrating excellent performance. Then, Xu et al. \cite{samamba} proposed SAMamba to efficiently deal with domain difference, which integrates hierarchical learning in SAM and global context modeling in Mamba. Jia et al. \cite{irgraphseg} proposed the IRGraphSeg based on graph deep learning, which can extract both image texture and structural information. Du et al. \cite{istd-diff} treated IRSTD as a mask-generation task and designed a conditional diffusion framework. Shi et al.\cite{AuxDet} achieved omni-domain infrared small target detection by incorporating auxiliary metadata.
    
    Although the detection performance is further improved, these methods rely on heavy computation cost with lots of parameters. More efficient methods with less hardware requirement and faster inference speed are expected.
	
	\subsection{Lightweight Networks for IRSTD}
	To achieve robust IRST detection with higher inference speed, many lightweight detection methods have been proposed. 
    
	The first kind of lightweight method is  model compression. These methods can be divided into the following categories. 
	
	\begin{itemize}
	\item[1)] Knowledge distillation \cite{14-MDPNet,15-IRKD}, which improves the performance of compact student model based on the knowledge of the cumbersome teacher networks. 
	\item[2)] Quantization \cite{16-SPMix-Q}, which uses lower bitwidth to represent network weight and activation. 
	\item[3)] Pruning \cite{17-SGBN,18-IRPruneDet}, which focuses on searching redundancy structure and then eliminating unimportant weights and channels. 
	\item[4)] Network architecture search \cite{19-darts,20-Bix-NAS}, which automatically explores the optimal network structure or component structure in the search space, combining with constraints and search strategies. 
	\item[5)] Low-rank factorization \cite{21-low-rank}, which uses the low-rank constraints of network to decrease the computation.
	\end{itemize}
    
	The second kind of method focuses on designing lightweight network structure \cite{22-ghostnet,23-vanillanet,24-LW-IRSTNet,25-RepISD-Net,26-SpirDet}, including the following.
	
	\begin{itemize}
	\item[1)] Reducing backbone depth, width, kernel size and improving accuracy with simple attention. 
	\item[2)] Replacing traditional convolution with lightweight convolution operators, such as separable convolution, group convolution and sparse convolution. 
	\item[3)] Accelerating model inference through re-parameterization. These methods heavily rely  on manual design and are therefore not as general as model compression methods.
	\end{itemize}

	Although the computational cost is further reduced, existing lightweight methods require lots of empirical knowledge and manual selection. More simple yet effective lightweight detection methods are expected to be explored.

	\subsection{Loss Functions for IRSTD}
    Appropriate loss function can accelerate the network convergence and increase the detection performance. Existing loss functions of small target detection field can be divided into three categories as follows.
    
    \begin{itemize}
    \item[1)] Distribution-based loss aims to minimize the distribution difference between predictions and ground truth, including cross-entropy loss \cite{CE}, Focal loss \cite{focal}. 
    \item[2)] Region-based loss aims to maximize the overlap area between predictions and masks, including Dice loss \cite{GDice}, SoftIoU loss \cite{softiou}. 
    \item[3)] Composite loss combines multiple types of losses, each of which complements the strengths of the others, including weighting, power and exponential.
	\end{itemize}
    
    Considering that IRSTD has highly unbalanced positive and negative samples, researches have designed many composite loss functions to achieve better and quicker network convergence. Zhang et al. \cite{13-isnet} used Dice loss and Edge loss to achieve accurate prediction of the target edge and shape. Edge loss (i.e., BCE loss + Dice loss) was designed to strengthen the utilization of target edge information. In space-based ship detection, Wu et al. \cite{32-mtu-Net}  combined the advantages of Focal loss and SoftIoU loss to strengthen target localization and shape description. Chen et al.\cite{DiceFocal} combined Dice loss and Focal loss to jointly optimize detection and segmentation tasks. Focusing on lack of sensitivity to target scale and position, Liu et al. \cite{35-SLSLoss} developed a novel scale and location sensitive (SLS) loss, enabling existing detectors to improve detection performance.
    
    \begin{figure*}
    	\centering
    	\includegraphics[width=18.2cm]{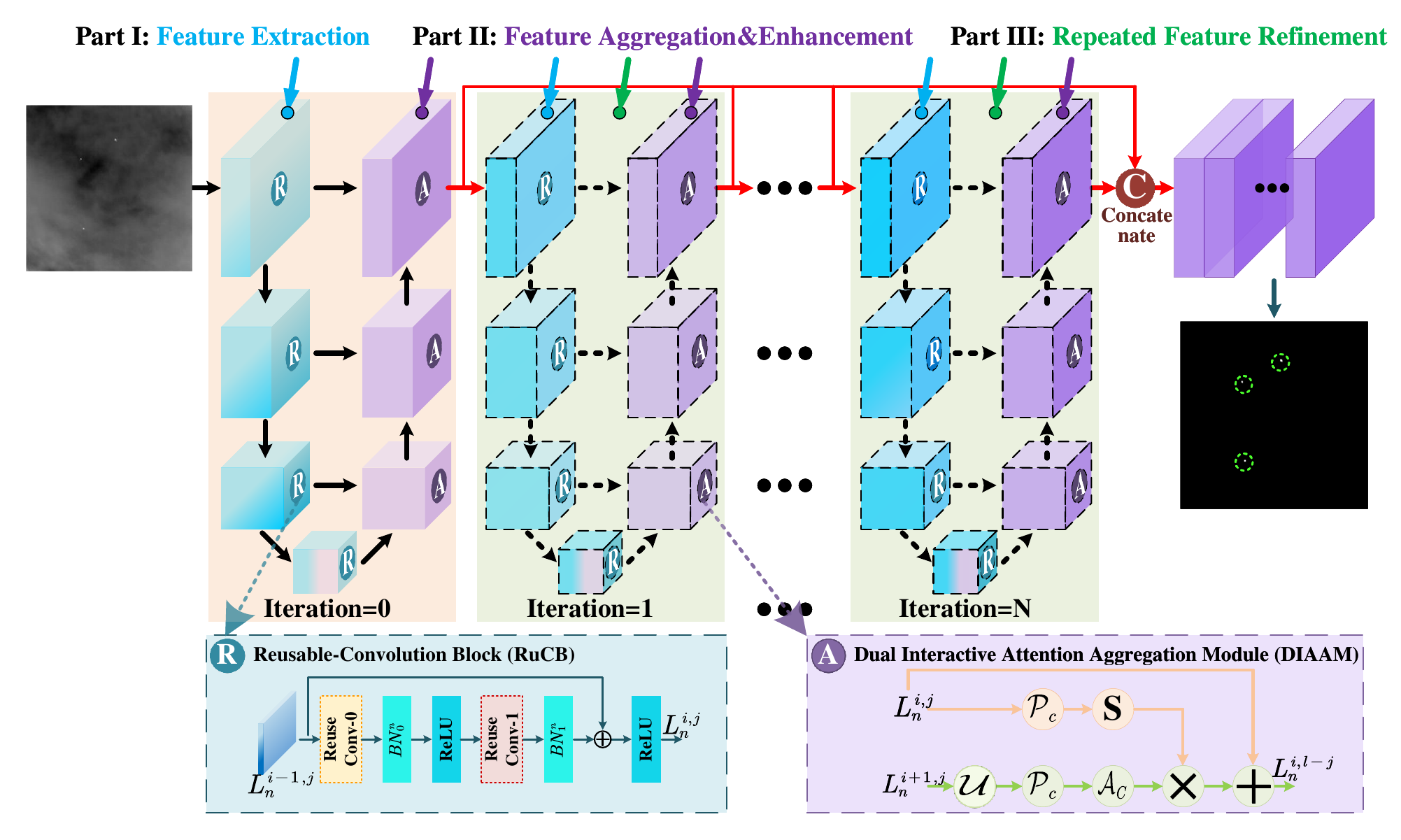}
    	\caption{An illustration of the proposed RRCA-Net. \textbf{Feature extraction}: Images are first input into the RuCB of encoder to extract multi-layer features. \textbf{Feature aggregation and enhancement}: Subsequently, in the decoder, the extracted features are upsampled and fused with a DIAAM. \textbf{Repeated feature refinement}: The multi-layer features are concatenated to achieve robust output refinement after several encoding and decoding iterations. Finally, eight-connected neighborhood clustering algorithm clusters the segmentation map to locate the centroid of each target region. }\label{network}
    \end{figure*}
	
	\section{Methodology}\label{SecMethodology}
	
	In this section, we introduce the RRCA-Net and its components in detail.
	
	\subsection{Overall Architecture}\label{SecOverall}
	
	As illustrated in Fig.~\ref{network}, our RRCA-Net takes a single infrared small target image as its input. Then, it sequentially executes the recurrent convolutional operation (Section \ref{SecRuCB}) and the simplified decoder (Section \ref{SecDIAAM}) to generate the pixel-level segmentation masks.
	
	Section \ref{SecRuCB} introduces the motivation of our recurrent structure and the architecture of the reusable-convolution block (RuCB). Input images are first preprocessed by a residual block and fed into the encoder backbone of RuCB to extract multi-scale features. Then, multi-scale features in different layers are gradually fused at the simplified asymptotic decoder, in which the dual interactive attention aggregation module (DIAAM) merges multi-level features to locate targets layer by layer. Next, the feature maps through the whole network are fed into the encoder again to refine deep feature iteratively, which constructs the recurrent structure. With the help of RuCB, high-level features are repeatedly refined with a  reusable convolution kernel without extra parameters.
	
	Section \ref{SecDIAAM} shows the strategy of feature fusion with a small amount of parameters. Due to the semantic gap at multi-scale feature fusion stage, we utilized DIAAM to adaptively enhance low-level and high-level features for performing well on interactive feature aggregation. Then, the shallow-layer features with rich spatial information and deep-layer features with rich semantic information are combined to enhance the informative feature maps. 
	
	Section \ref{SecFSM} introduces the FSM module that aggregates fine-grained features from multiple iterations. The robust outputs are fed into the eight-connected domain clustering module \cite{12-DNANet} to calculate intersection over union, probability of detection and false alarm rate, which are then used for comparison in Section \ref{SecExperiment}.

	\subsection{The Recurrent Structure with Iterations}\label{SecRuCB}
	
	\subsubsection{Motivation}
	
	The general U-shape structure consists of an encoder, a decoder, and plain skip connections. The encoder is used to extract multi-level features and decoder helps to achieve progressive feature fusion. To achieve powerful modeling capability, a straightforward way is to continuously increase the number of convolution layers, including dense skip connection \cite{12-DNANet}, nesting U-Net inside U-Net \cite{13-UIUNet}, cascading several U-Net \cite{doubleunet} and so on. However, these methods introduces huge computation cost. Therefore, we try to design a simple structure to achieve similar performance with fewer parameters and less computation cost.

	\subsubsection{The Reusable-Convolution Block in Encoder}
	As shown in Fig.~\ref{network}, we extract the features from the first encoding-decoding iteration. After feature aggregation and enhancement, these features are fed into the reusable encoder again with N-time iteration.  The reusable operation can make the network deeper without inserting additional new convolution layers. Moreover, convolution kernels can get gradient update from all re-usable operations. This helps the RRCA-Net to precisely capture informative feature details of small targets.
	
	Specifically, RuCB reuses \textit{N} $3\times3$ convolutions in iterations. It is worth noting that, we only repeatedly use the convolution operation but set new batch normalization (BN) layer for each reusing block since the output feature from different iteration has  different data distribution. BN needs to learn new scale and shift parameters to accelerate the train process. Therefore, recurrent iteration will introduce a slight increase in parameters.
	
	In this paper, we stack $N$ RuCBs to form our recurrent structure. Without loss of generality, we take the RuCB in $n^{th}(n=0,1,2,...,N)$ recurrent iteration as an example to explain our method. As shown in Fig.~\ref{network}, Assume $\mathbf{L}_{n}^{i,j}$ denotes the output of node $\mathbf{L}(i,j)$, where $i$ is the $i^{th}$ layer in the encoder. $j$ is the $j^{th}$ convolution layer along the plain skip connection. $n$ is the $n^{th}$ iteration in recurrent structure. When $n=0$, the encoder only use plain residual convolution block without iteration. The processed features maps represented by $\mathbf{L}_{n}^{i,j}$ are calculated as follows.
	
	\begin{equation}\label{DNIM_1}
		{\mathbf{L}}_{n}^{i,j}={\mathcal{P}}_{max}({\mathcal{F}}(\mathbf{L}_{n}^{i-1,j})),
	\end{equation} where $\mathcal{F}(\cdot)$ denotes multiple cascaded residual convolution block. For each layer from 0 to 3, the corresponding number of cascaded block is (3,2,2,2). ${\mathcal{P}}_{max}(\cdot)$ denotes max-pooling with a stride of 2.
	
	When $n>0$ , each node in encoder utilizes RuCB to iterate recurrently. The processed feature maps represented by $\mathbf{L}_{n}^{i,j}$ is generated as follows.
	
	\begin{equation}\label{DNIM_2}
		{\mathbf{L}}_{n}^{\textit{i},\textit{j}}={\cal{P}}_{max}({\cal{R}}(\mathbf{L}_{n}^{i-1,j})),
	\end{equation} where $\cal{R}(\cdot)$ denotes the cascaded reusable-convolution block.
	
	\subsection{The Dual Interactive Attention Aggregation Module}\label{SecDIAAM}
		
	After feature extraction in the encoder, we develop a dual interactive attention aggregation module to fuse the multi-layer features. As shown in Fig.~\ref{network}, the high-level and shallow-level features of the adjacent layers are input into the DIAAM at the same time, so that the information in multi-layer features is gradually aggregated. 
	
	Specially, the high-level features $\mathbf{L}_{n}^{i+1,j}$ are first amplified by a factor of two through bilinear upsampling and its channels are cut in half by point-wise convolution. In this way, the high-level and shallow-level feature can keep the same dimension. Next, channel-wise attention enhances the responses of those features located on informative feature channels. Moreover, in another branch, the shallow-level feature $\mathbf{L}_{n}^{i,j}$ are processed to generate attention. By weighting $\mathbf{L}_{n}^{i,j}$ to $\mathbf{L}_{n}^{i+1,j}$, i.e. attention mechanism, the detailed information of the previous layer is used to highlight the target information in high-level features and suppress noises. Finally, the processed features and the shallow-level features are element-wisely summed. In this way, DIAAM helps to achieve feature fusion between shallow-level feature with full spatial and profile information and high-level feature with rich semantic information.

	Let's assume that ${\mathbf{L}}_{n}^{i,l-i}$ denotes the output of node ${\mathbf{L}}(i,l-i)$ in decoder. $l$ is the depth of network. The formula for DIAAM is given as follows.  
	
	\begin{equation}\label{DNIM_2}
		{\mathbf{L}}_{n}^{i,l-j}={\mathbf{L}}_{n}^{i,j}+ \mathcal{S}({{\mathcal{P}}_{c}}({\mathbf{L}}_{n}^{i,j})) \otimes  \mathcal{A}_{c}({\mathcal{P}}_{c}({\mathcal{U}}(\mathbf{L}_{n}^{i+1,j}))),
	\end{equation} where $\mathcal{U}(\cdot)$ denotes the upsampling operation based on bilinear interpolation.  $\mathcal{S}(\cdot)$ denotes the Sigmoid activation function. ${\mathcal{P}}_{c}(\cdot)$ denotes the point-wise convolution. $\otimes$ represents the element-wise multiplication. $\mathcal{A}_{c}(\cdot)$ denotes the channel attention mechanism, which is calculated as:
	
	\begin{equation}\label{SCAM_1}
		\mathbf{F}^{'}=\mathbf{F} \otimes {\mathcal{S}}(MLP({\mathcal{P}}_{max}(\mathbf{F})) + MLP({\cal{P}}_{avg}(\mathbf{F}))) ,
	\end{equation} where $\mathbf{F}$ is the input feature maps and $\mathbf{F}^{'}$ is the processed output. $\mathcal{P}_{avg}(\cdot)$ denotes the average pooling with a stride of 2. $MLP(\cdot)$ denotes the shared multi-layer perceptron.

	\subsection{The Feature Stacking Module}\label{SecFSM}
	
	Through $N$ iterations in the recurrent structure, we will obtain $N$ feature maps from each iteration. In order to make full use of features across multiple granularities, we stack these features together  and output predictions. As shown in Fig.~\ref{network}, these decoded features from different iterations are concatenated in channel dimension and fed into the last stage block. Finally, both coarse-grained and fine-grained features are merged to generate robust masks. The strategy is stated as:

	\begin{equation}\label{FSM}
		\mathbf{O}_{FSM}=\left[\mathbf{L}^{0,l}_{0},\mathbf{L}^{0,l}_{1},...,\mathbf{L}^{0,l}_{N}\right],
	\end{equation} where $\left[\cdot,\cdot \right]$ represents the concatenate operation.

	\subsection{The Eight-connected Domain Clustering Algorithm for Evaluation}\label{SecCluster}
	
	After obtaining the output, we introduce an eight-connected domain clustering algorithm to calculate $P_d$ and $F_a$. In this way, we can cluster the pixels within the same connected domain and then compute the centroid coordinates of each target. In feature maps $\mathbf{G}$, when a pixel ${(x_{0},y_{0})}$ is in the eight-neighborhood of another pixel ${(x_{1},y_{1})}$ and these two pixels have the same value (0 or 1), they are judged to belong to the same connected domain, i.e.,

	\begin{equation}\label{ENCM_3}
		\mathbf{g}_{(x_{0},y_{0})}=\mathbf{g}_{(x_{1},y_{1})}, \exists \left( x_0,y_0 \right) \in \mathcal{N}_8\left( x_1,y_1 \right) ,
	\end{equation} where ${{\cal{N}}_{8}}_{(x_{1},y_{1})}$ represents the eight-neighborhood of pixel ${(x_{1},y_{1})}$, $\mathbf{g}_{(x_{0},y_{0})}$ and $\mathbf{g}_{(x_{1},y_{1})}$ are the gray value of pixel  ${(x_{0},y_{0})}$ and ${(x_{1},y_{1})}$. Pixels in a connected domain belong to the same target. Then centroids can be calculated according to the coordinates of all pixels in targets.
	
	\subsection{DpT-k Loss for Infrared Small Target Segmentation}\label{DpT-kLoss}
	
	With the development of IRSTD, researchers have demonstrated that appropriate loss functions can help networks to improve performance and accelerate training process. Dice loss \cite{GDice} is derived from the dice similarity coefficient, which is widely used in medical image segmentation. Dice loss can describe regional correlation and are sensitive to target scale, but it is prone to lose small-scale targets and edge pixels. Assume that $P$ and $G$ are the prediction and ground truth of image $I$, respectively. Dice loss is defined by:
	
	\begin{equation}\label{DiceLoss}
		\mathcal{L}_{Dice}=1-\frac{2\sum_{c=1}^C{\sum_{i=1}^V{p_{i}^{c}}}g_{i}^{c}}{\sum_{c=1}^C{\sum_{i=1}^V{p_{i}^{c}}}+\sum_{c=1}^C{\sum_{i=1}^V{g_{i}^{c}}}},
	\end{equation} where $p_i$, $g_i$ respectively denote the values of pixel $i$ in $P$, $G$. $V$ is the number of pixels in image $I$. $C$ is the number of classes, and $C=2$ since IRSTD is a binary classification problem.

	Top-k loss \cite{topkloss} and Poly loss \cite{polyloss} are the variants of CE loss, which are used to measure distributions and are inferior to match shape. Poly loss analyzes CE loss from polynomial expansion perspective, and tunes the first polynomial term to obtain a significant gain. It is defined by:
	
	\begin{equation}\label{PolyLoss}
		\mathcal{L}_{Poly}=-\frac1V  \sum_{c=1}^C\sum_{i=1}^V{g_i^c}\mathrm{log}({p_i^c}) + \alpha(1-\frac1V\sum_{c=1}^C\sum_{i=1}^V{g_i^c}{p_i^c}),
	\end{equation} where $\alpha$ is the perturbation to adjust the contribution of the first term.
	
	\begin{figure}
		\centering
		\includegraphics[width=8.8cm]{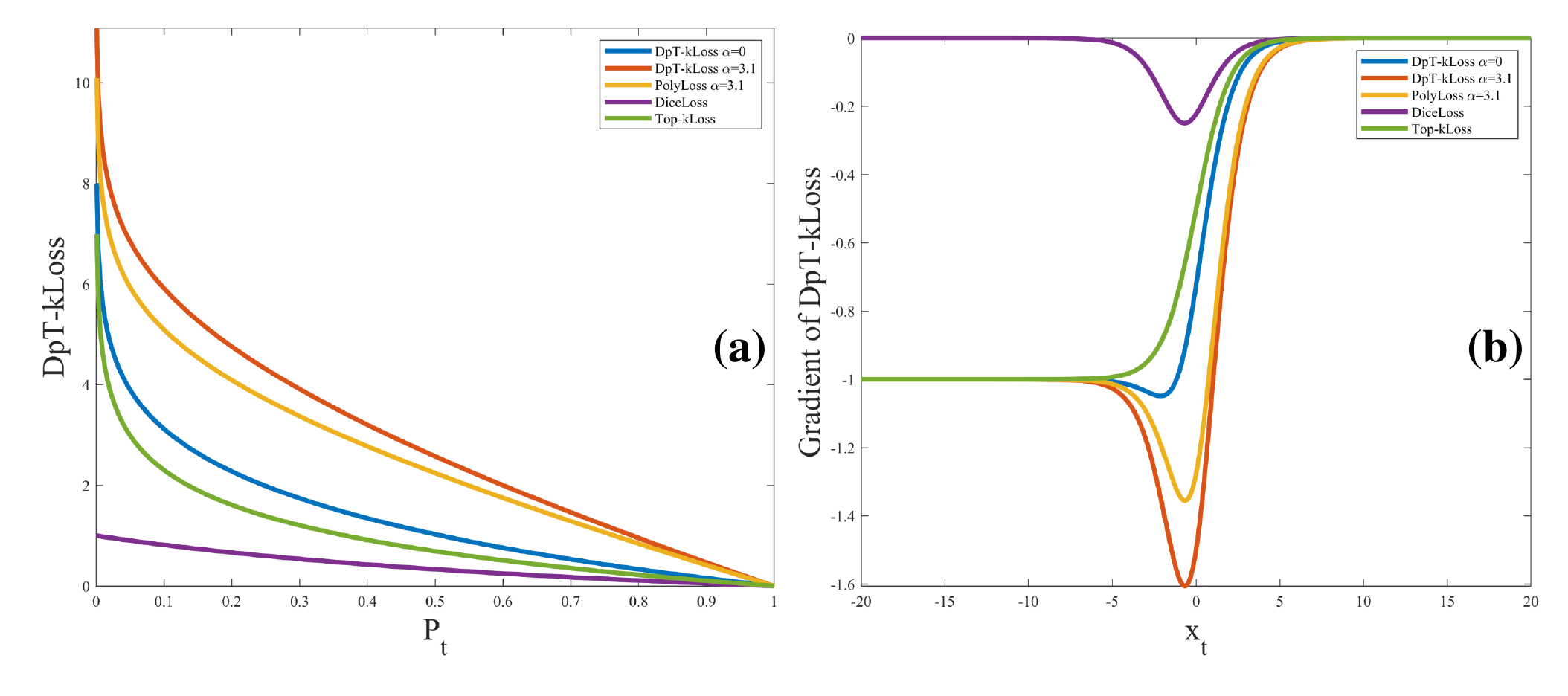}
		\caption{Visual analysis of DpT-k loss and others. (a) Curves of loss. (b) Gradient curves of loss.}\label{losscurve}
	\end{figure}
	
	Top-k loss retains top $k\%$ of the worst predicted pixel for calculating values, which aims to force network to focus on hard samples and adapts to imbalanced distributions. In other words, the ranking criterion is the value of cross entropy between predicted pixels and groundtruth pixels. The value of Top-k is calculated by the largest k\% values of all cross entropy between pixels. It is defined by:
	
	\begin{equation}\label{TopkLoss}
		\mathcal{L}_{Top\_k}=-\frac1V \sum_{c=1}^C\sum_{i\in\mathbf{K}}^V{g_i^c}\mathrm{log}({p_i^c}),
	\end{equation} where $\mathbf{K}$ is the set of $k\%$ worst predicted pixels.
	
	Combining the advantages of both classes of loss functions, we propose the composite loss with shape and location sensitivity. The formula is defined by:
	
	\begin{equation}\label{DpTkLoss}
		\mathcal{L}_{DpT\_k}=\mathcal{L}_{Dice} + \mathcal{L}_{Poly} + \mathcal{L}_{Top_k}.
	\end{equation}
	
	As shown in Fig.~\ref{losscurve}, we can find that DpT-k loss gives a higher loss value when the prediction probability is same. And as $p$ increases, DpT-k loss decreases more sharply. Subsequent experiments show that DpT-k loss introduces large gradients to negative samples, which helps network to suppress false alarms. DpT-k loss focuses on hard samples that are difficult to learn, the influce of already learned targets is weakened and overfitting is avoided. With shape and distribution constrains, the DpT-k loss can both refine the contour description and achieve accurate centroid localization, which supports performance improvement of the network.

	\section{Experiment}\label{SecExperiment}
	In this section, we first introduce our evaluation metrics and implementation details. Then, comparison experiments and ablation studies are presented.
	
	\subsection{Evaluation Metrics}\label{Evaluation Metrics}
	
	Detection performance metrics include intersection over union (IoU), probability of detection (${P}_{d}$), false alarm rate (${F}_{a}$) and receiver operating characteristic (ROC) curves.
    
    Computation cost metrics include $\#$Params and FLOPs.

	\begin{table*}[t]
		\centering
		\renewcommand\arraystretch{1.2}
		\caption{Results comparison of detection performance ($IoU$($\times10^{-2}$), $P_{d}$($\times10^{-2}$), and $F_{a}$($\times10^{-6}$)) and model size ($\#Params$(M)) achieved by different state-of-the-art methods on three datasets. The best results are in \textcolor{red} {red} and the second best results are in \textcolor{blue} {blue}.} \label{Tabcomparisonmethod}
		\begin{tabular}{l c c c c c c c c c c c}
			\hline
			\multicolumn{1}{c}{\multirow{2}{*}{Method   Description}} & \multicolumn{3}{c}{IRSTD-1k\cite{13-isnet}}         & \multicolumn{3}{c}{NUAA-SIRST\cite{11-ACM}} & \multicolumn{3}{c}{DenseSIRST\cite{DenseSIRST}}  & \multicolumn{1}{c}{\multirow{2}{*}{$\#$Params$\downarrow$}} & \multicolumn{1}{c}{\multirow{2}{*}{$\#$FLOPs$\downarrow$}} \\ \cline{2-10}
			\multicolumn{1}{c}{}   & \multicolumn{1}{c}{$IoU\uparrow$} & \multicolumn{1}{c}{$P_{d}\uparrow$} & \multicolumn{1}{c}{$F_{a}\downarrow$} & \multicolumn{1}{c}{$IoU\uparrow$} & \multicolumn{1}{c}{$P_{d}\uparrow$} & \multicolumn{1}{c}{$F_{a}\downarrow$} & \multicolumn{1}{c}{$IoU\uparrow$} & \multicolumn{1}{c}{$P_{d}\uparrow$} & \multicolumn{1}{c}{$F_{a}\downarrow$} & \multicolumn{1}{c}{} & \multicolumn{1}{c}{}                \\ \hline
			Top-Hat \cite{5-tophat}     &   10.06    &  75.11     & 1432  & 7.14  & 79.84  & 1012 & 1.07 & 32.77 & 563.9 & - & - \\
			Max-Median \cite{6-maxmedian}  & 7.00 & 65.21 & 59.73 & 4.17 & 69.20 & 55.33 & 0.12 & 33.24 & 5.178 &- & - \\ 
			WSLCM \cite{7-WSLCM} & 3.45 & 72.44 & 6619 & 1.16 & 77.95 & 5446 & 2.58 & 48.90 & 2100 & - & -      \\  
			TLLCM \cite{8-TLLCM}        & 3.31  & 77.39 & 6738 & 1.03 & 79.09 & 5899 & 1.12 & 22.15 & 2300 &- & -\\ 
			IPI \cite{8-IPI}     & 27.92 & 81.37 & 16.18 & 25.67 & 85.55 & 11.47 & 3.35 & 42.13 & 10.45 & - & -  \\ 
			NRAM \cite{10-NRAM}  & 15.25 & 70.68 & 16.93 & 12.16 & 74.52 & 13.85 & 4.91 & 65.32 & 4.363 & - & -    \\ 
			RIPT \cite{10-RIPT}  & 14.11 & 77.55 & 28.31 &  11.05  & 79.08 & 22.61 & 5.98 & 72.93 & 9.179 & - & -\\ 
			PSTNN \cite{9-PSTNN} & 24.57 & 71.99 & 35.26 & 22.40  & 77.95  & 29.11 & 4.77 & 35.12 & 138.7 & - & - \\ 
			MSLSTIPT \cite{4-anti-miss} & 11.43 & 79.03 & 1524 & 10.30 & 82.13 & 1131 & 2.13 & 17.37 & 333.2 &- & -\\ \hline
			ACM \cite{11-ACM} & 60.33 & 92.27 & 65.31 & 68.34 & 91.63 & 23.19 & 53.64 & 61.75 & 31.56 & 0.398 & \textcolor{blue}{0.335} \\ 
			ResUNet \cite{resunet}& 66.06 & 90.48 & 10.63 & 74.06 & 96.96 & 14.97 & 80.96 & 98.06 & 4.377 & 0.914 & 1.27 \\
			AMFU-net \cite{AMFU}   & 66.70 & \textcolor{red}{92.86} & 16.02 & 74.26 & 97.34 & 14.40 & 80.33 & 97.89 & 4.132 & 0.473 & 5.09 \\
			DNANet \cite{12-DNANet} & \textcolor{blue} {67.70} & 91.58 & 18.69  & \textcolor{red}{76.63} & 95.44 & 26.55 & 81.45 & 97.28 & 3.252 & 4.697 & 14.05\\ 
			UIU-Net \cite{13-UIUNet} & 65.69 & 91.25 & 13.42 & \textcolor{blue}{76.24} &  95.06 & 10.57 & 77.51 & 97.11 & 6.699 & 50.54 & 46.65  \\ 
			LW-IRSTNet \cite{24-LW-IRSTNet} & 61.34 &  89.46  & 23.84 & 67.83 & 94.30  & 19.39 & 78.95 & 97.11 & 3.656 & \textcolor{blue}{0.163} & \textcolor{red}{0.177} \\ 
			RepISD-Net  \cite{25-RepISD-Net} & 65.45 & 91.59 & 7.62 & 74.87  & \textcolor{red}{98.48} & \textcolor{red}{6.85} & 78.60 & 97.25 & 22.24 & 0.279 & 11.53 \\ \hline
			RRCA-Net-ResNet10 (ours) & 66.57 & 90.48 & 13.36  & 74.17  & 96.20  & 20.25 & \textcolor{red}{82.35} & 98.03 & \textcolor{red}{1.319} & \textcolor{red}{0.103} & 0.721 \\ 
			RRCA-Net-ResNet18 (ours) & \textcolor{red} {68.73}  & \textcolor{blue}{92.52} &\textcolor{red} {6.68}  & 75.41 & \textcolor{blue} {98.10}  & \textcolor{blue} {7.56} & \textcolor{blue}{82.15} & \textcolor{blue}{98.23} & \textcolor{blue}{1.810} & 0.202 & 1.04 \\ 
			RRCA-Net-ResNet34 (ours)  & 67.31  & \textcolor{blue}{92.52}  & \textcolor{blue}{7.29}  & 72.45  & 96.96   &  12.34 & 82.00 & \textcolor{red}{98.49} & 2.308 & 0.363 & 1.68 \\ \hline
		\end{tabular}
	\end{table*}
	
	\subsubsection{Intersection over Union}
	Intersection over Union ($IoU$) is a pixel-level evaluation metric, which mainly reflects the profile description ability of the algorithm. IoU is numerically equal to the ratio of the  intersection and the union between the predictions and labels, i.e.,
	{\begin{equation}\label{mIoU}
			{IoU}= \frac{Area_{inter}}{Area_{Union}},
		\end{equation}where $Area_{inter}$ and $Area_{Union}$ represent the interaction areas and union areas between the predictions and labels, respectively.}

	\subsubsection{Probability of Detection}
	Probability of Detection (${P}_{d}$) is a target-level evaluation metric, which mainly reflects the target localization accuracy of the algorithm. ${P}_{d}$ measures the ratio of correctly predicted target number $Target_{correct}$ over all target number $Target_{All}$.  ${P}_{d}$ is defined as follows:
	
	{\begin{equation}\label{PD}
			{P}_{d}= \frac{Target_{correct}}{Target_{All}},
	\end{equation}}
	
	If the centroid deviation of the target is less than the predefined threshold $ D_{\textit{thresh}}$, it's considered those targets as correctly predicted ones. We set the deviation threshold as 3 in this paper.

	\subsubsection{False Alarm Rate}
	False Alarm Rate (${F}_{a}$) is another target-level evaluation metric, which mainly reflects the wrong detection of false targets. ${F}_{a}$ is equal to the ratio of falsely predicted pixels $Pixel_{false}$ over all image pixels $Pixel_{All}$. ${F}_{a}$ is defined as follows:
	{\begin{equation}\label{FA}
			{F}_{a} = \frac{Pixel_{false}}{Pixel_{All}},
	\end{equation}}

	If the centroid deviation of the target is larger than the predefined threshold, we consider those target pixels as falsely predicted ones. We set the predefined deviation threshold as 3 in this paper.

	\subsubsection{Receiver Operation Characteristics}
	
	Receiver Operation Characteristics (ROC) is used to describe the shifting trends of the detection probability (${P}_{d}$) under varying false alarm rate (${F}_{a}$). ROC indicates the total effect under a sliding threshold, comprehensively reflecting the overall performance of the network on ${P}_{d}$ and ${F}_{a}$.
	
	\subsubsection{$\#$Params}
	Parameters refers to the total number of learnable parameters in the model, which reflects the model size. Taking the conventional 2D convolution as an example, input feature map is $F\in \mathbb{R}^{C_{in}\times H_{in}\times W_{in}}$, 2D convolution kernel is $C\in \mathbb{R}^{C_{in}\times K_{h}\times K_{w}}$ and output feature map is $O\in \mathbb{R}^{C_{out}\times H_{out}\times W_{out}}$. $\#Params$ is related to the size of convolution kernel. The calculation formula is
	{\begin{equation}\label{Params}
			\#Params=K_h\times K_w\times C_{in}\times C_{out},
	\end{equation}}where $C_{out},\ C_{in},\ K_h,\ K_w$ represent the output channel number, input channel number, convolution kernel height and convolution kernel width of 2D convolution, respectively.
	\subsubsection{FLOPs}
	 FLOPs refer to the total number of float point operations performed during the model's execution, including addition, subtraction, multiplication and division operations. Taking the conventional 2D convolution as an example, FLOPs are related to the input feature and convolution kernels. The calculation formula is 
 	{\begin{equation}\label{FLOPs}
 		FLOPs=2\times K_h\times K_w\times C_{in}\times C_{out}\times H_{in} \times W_{in}.
 		\end{equation}}

	\subsection{Implementation Details}
	We used three benchmark datasets, including NUAA-SIRST\cite{11-ACM}, IRSTD-1k\cite{13-isnet} and DenseSIRST dataset \cite{DenseSIRST} for quantitative experiments, which respectively consist of 427, 1001, and 1024 images. We followed  the previous methods in \cite{12-DNANet} and keep the same setting for NUAA-SIRST. IRSTD-1k and DenseSIRST datasets. Before training, all input images are first normalized and sequentially processed by random image flip and crop for data augmentation. Finally, these images have a resolution of $256 \times 256$ before being fed into the network.
	
	The backbone of RRCA-Net is the typical U-Net paradigm with ResNet \cite{Resnet} and the network is initialized using the Xavier strategy \cite{Xavier}. It is trained using the DpT-k Loss ($k=10$, $\alpha=3.1$) and optimized by the AdaGrad method \cite{Adagrad} with the CosineAnnealingLR scheduler \cite{CosineLR}. We set the learning rate, batch size, epoch number as 0.05, 8 and 1500, respectively. The traditional detection methods are implemented in MATLAB on an Intel Core i9-13900HX CPU, while those methods based on deep learning are trained and tested in PyTorch \cite{pytorch} on an NVIDIA GeForce RTX 3090 GPU.

	\subsection{Comparison to the State-of-the-art Methods}\label{SOAT}
	To demonstrate the effectiveness of our method, we compare our RRCA-Net to several SOTA IRSTD methods, including \textbf{traditional methods:} (filtering-based methods: Top-Hat\cite{5-tophat} and  Max-Median \cite{6-maxmedian}; local-contrast-based methods: WSLCM \cite{7-WSLCM} and TLLCM\cite{8-TLLCM}; low-rank-based methods: IPI\cite{8-IPI}, PSTNN\cite{9-PSTNN}, NRAM\cite{10-NRAM}, RIPT\cite{10-RIPT} and MSLSTIPT \cite{4-anti-miss}) and \textbf{CNN-based methods:} (ACM\cite{11-ACM}, ResUNet\cite{resunet}, AMFU-net\cite{AMFU},  DNANet\cite{12-DNANet}, UIU-Net\cite{13-UIUNet}, LW-IRSTNet \cite{24-LW-IRSTNet} and RepISD-Net \cite{25-RepISD-Net}) on the NUAA-SIRST, IRSTD-1k and DenseSIRST datasets. For fair comparison, we collect and implement the official open-source code of those methods. Then, we keep the same training settings of their original papers and retrain all the CNN-based methods on the same training datasets.

	\subsubsection{Quantitative Results}
	For all the compared algorithms, we first obtain their predictions and then suppress low-response noise by setting a threshold to calculate evaluation metrics of detection performance. Traditional methods involve calculating the corresponding adaptive threshold. For CNN-based methods, we follow their original papers and adopted their fixed thresholds. We keep all remaining hyper-parameters the same as their original papers.

	Quantitative results are shown in Table~\ref{Tabcomparisonmethod}. The deep-learning-based methods significantly outperform the traditional algorithms in terms of both detection accuracy (i.e., PD) and contour description (i.e., IoU). Limited by handcraft features selection, those model-driven traditional methods have difficulty in coping with complex and varied scenes. Meanwhile, Our RRCA-Net achieves competitive performance at a low computational cost, particularly attaining the highest IoU value and the lowest ${F}_{a}$ value compared to other methods on IRSTD-1k dataset.

	\begin{figure*}[htbp]
		\centering
		\subfigure[]{
			\begin{minipage}[t]{0.33\linewidth}
				\centering
				\includegraphics[width=6.0 cm]{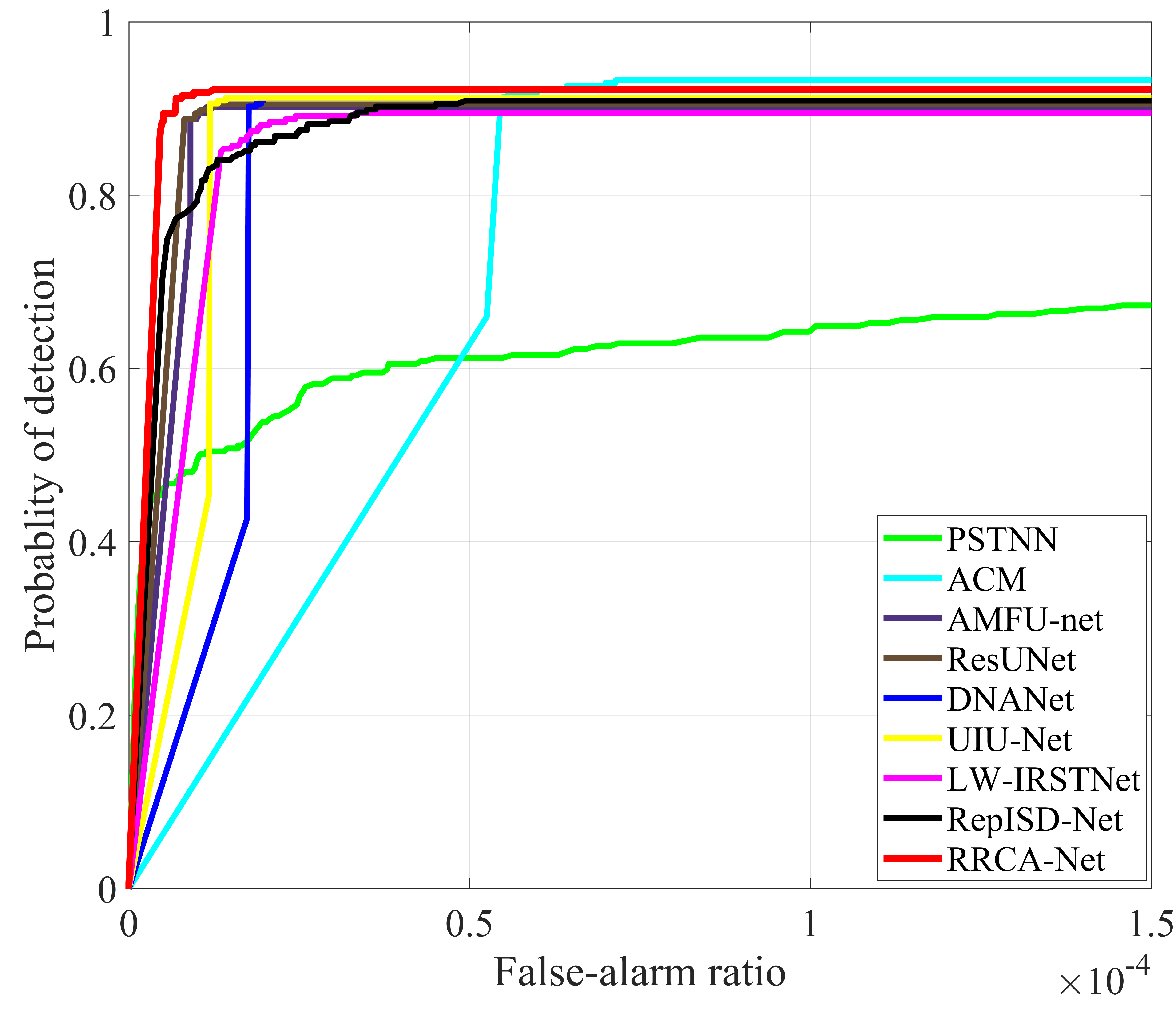}
			\end{minipage}%
		}%
		\subfigure[]{
			\begin{minipage}[t]{0.33\linewidth}
				\centering
				\includegraphics[width=6.0 cm]{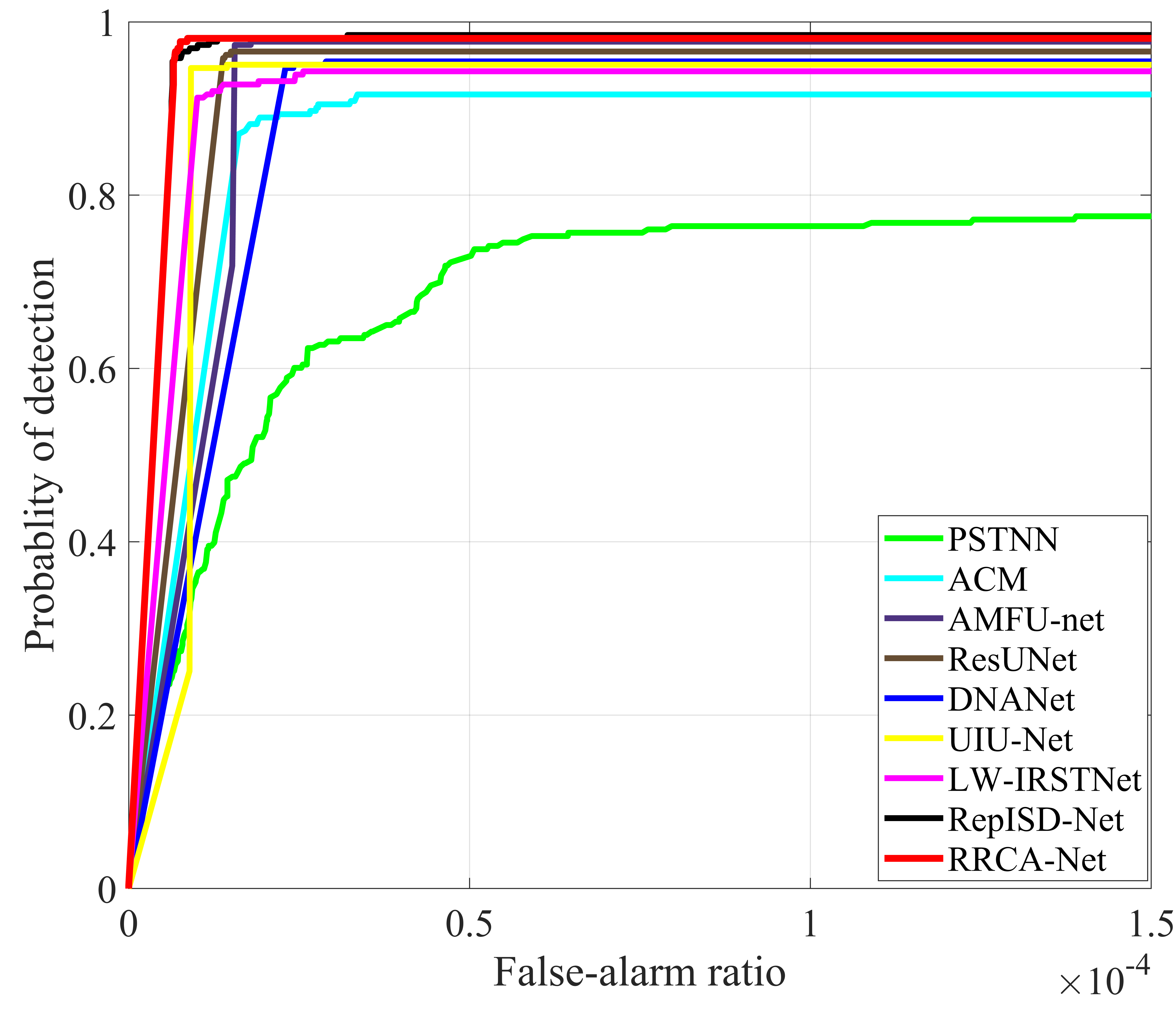}
			\end{minipage}%
		}%
		\subfigure[]{
			\begin{minipage}[t]{0.33\linewidth}
				\centering
				\includegraphics[width=6.0 cm]{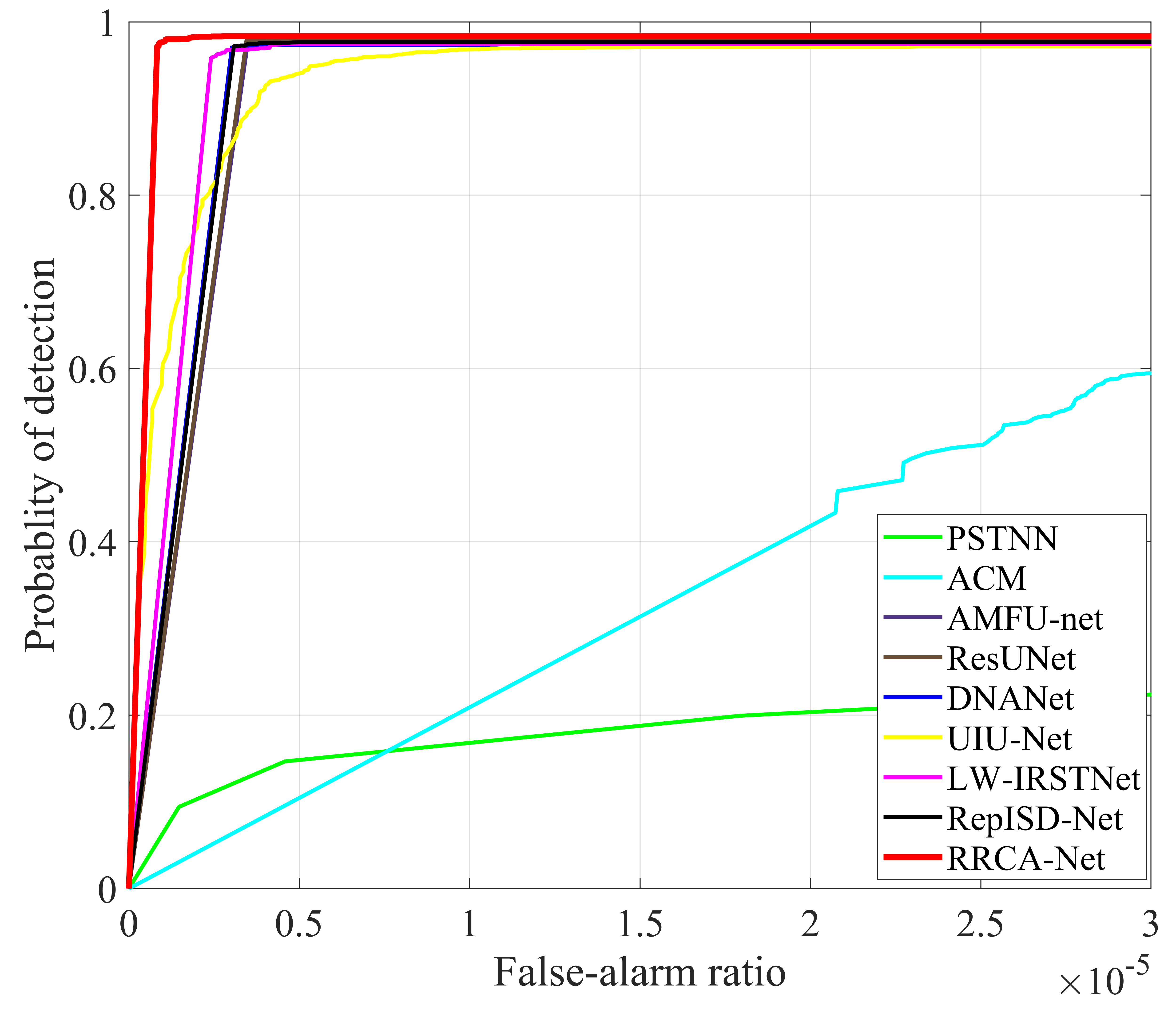}
			\end{minipage}%
		}%
		\caption{ROC performance of different methods on (a) IRSTD-1k, (b) NUAA-SIRST and (c) DenseSIRST datasets, respectively. Our RRCA-Net achieves consistent superior performance under different thresholds.}\label{ROCcurves}
	\end{figure*}

	\begin{table*}[h]
	\centering
	\renewcommand\arraystretch{1.2}
	\caption{Results comparison of CNN-based methods with RuCB. \textbf{Bold} indicates performance improvement after applying RuCB on the original networks.} \label{transfer}
	\begin{tabular}{l c c c c c c c c c c c}
		\hline
		\multicolumn{1}{c}{\multirow{2}{*}{Method   Description}} & \multicolumn{3}{c}{IRSTD-1k\cite{13-isnet}}         & \multicolumn{3}{c}{NUAA-SIRST\cite{11-ACM}} & \multicolumn{3}{c}{DenseSIRST\cite{DenseSIRST}}  & \multicolumn{1}{c}{\multirow{2}{*}{$\#$Params$\downarrow$}} & \multicolumn{1}{c}{\multirow{2}{*}{$\#$FLOPs$\downarrow$}} \\ \cline{2-10}
		\multicolumn{1}{c}{}   & \multicolumn{1}{c}{$IoU\uparrow$} & \multicolumn{1}{c}{$P_{d}\uparrow$} & \multicolumn{1}{c}{$F_{a}\downarrow$} & \multicolumn{1}{c}{$IoU\uparrow$} & \multicolumn{1}{c}{$P_{d}\uparrow$} & \multicolumn{1}{c}{$F_{a}\downarrow$} & \multicolumn{1}{c}{$IoU\uparrow$} & \multicolumn{1}{c}{$P_{d}\uparrow$} & \multicolumn{1}{c}{$F_{a}\downarrow$} & \multicolumn{1}{c}{} & \multicolumn{1}{c}{}               \\ \hline
		ResUNet \cite{resunet}    & 66.06 & 90.48 & 10.63 & 74.06 & 96.96 & 14.97 & 80.96 & 98.06 & 4.377 & 0.9141 & 1.27 \\
		AMFU-net \cite{AMFU}   & 66.70 & 92.86 & 16.02 & 74.26 & 97.34 & 14.40 & 80.33 & 97.89 & 4.132 & 0.4730 & 5.09 \\
		DNANet \cite{12-DNANet} &  67.70 & 91.58 & 18.69  & 76.63 & 95.44 & 26.55 & 81.45 & 97.28 & 3.252 & 4.697 & 14.05 \\ \hline
		ResUNet with RuCB & \textbf{67.81} & \textbf{92.86} & \textbf{9.489} & \textbf{75.13} & 95.82 & \textbf{8.842} & \textbf{83.12} & \textbf{98.58} & \textbf{1.096} & 0.979 & 1.68 \\ 
		AMFU with RuCB & 66.19 & 92.59 & \textbf{5.436} & \textbf{74.30} & 95.06 & \textbf{5.419} & \textbf{81.48} & 97.77 & \textbf{1.796} & 0.489 & 5.77\\
		DNANet with RuCB  & \textbf{69.29} & 91.50 & \textbf{5.162} & 75.84 & \textbf{96.58} & \textbf{5.134} & 80.80 & \textbf{98.12} & \textbf{2.394} & 4.697 & 15.04 \\ \hline
	\end{tabular}
	\end{table*}

	On NUAA-SIRST dataset, the DNANet and UIU-Net achieved the top two performance in term of IoU, they don't perform as well as RepISD-Net and our RRCA-Net on ${P}_{d}$ and ${F}_{a}$. RRCA-Net achieves the competitive detection performance to RepISD-Net in three metrics. RRCA-Net achieves a 75.41$\%$ performance in term of IoU, which is slightly lower than the highest value (i.e., DNANet) 1.22$\%$, but with 2.66$\%$ increase in ${P}_{d}$ and a significant decrease in term of ${F}_{a}$. On IRSTD-1k dataset, our RRCA-Net achieves the optimum performance with 68.73$\%$ in IoU, 92.52$\%$ in ${P}_{d}$ and only 6.68$\times10^{-6}$ in ${F}_{a}$. AMFU-net only slightly outperforms our method by 0.34$\%$ in ${P}_{d}$. That is because that IRSTD-1k dataset contains realistic images in large resolution and focuses on the annotation of target shape with accurate pixel-level masks, which raises the challenge of detecting edges. On the Dense-SIRST dataset, RRCA-Net achieves the best detection performance with low false alarms. The dataset contains numerous dense clusters of small targets, most of which are only a few pixels apart. Multiple iterations in RRCA-Net can accurately separate the close targets. In summary, RRCA-Net maintains details information and prevents semantic loss in deep layers through multiple iterative refinement of the recurrent structure. And DIAAM efficiently integrates low-level and high-level features in a mutually guided manner, thereby preventing from introducing noise interference and suppressing false alarms. 
	
	As for computational efficiency, RRCA-Net with ResNet-10 only has 0.103M parameters and still achieves a advanced detection performance under such low model parameters. That is because that reusing convolution does not introduce extra parameters, and can also effectively extract deep features. It helps to extract informative multi-level features after multiple feature refinement and thus introduces accurate detection. But FLOPs increase because of using multiple reusing to process features.
	
	Overall, our proposed RRCA-Net  achieves a balance between detection performance and model size.
	
	As shown in Fig.~\ref{ROCcurves}, we plot ROC curves of different methods on three datasets. It is evident that our RRCA-Net can achieve competitive performance to other methods on all datasets under low computational cost, especially on those challenging IRSTD-1k and DenseSIRST dataset, RRCA-Net can effectively detect changeable targets from complex background. It can be seen that RRCA-Net can more stably achieve high ${P}_{d}$ while maintaining low ${F}_{a}$ than the state-of-the-art methods.

	\begin{figure*}
	\centering
	\includegraphics[width=18.2cm]{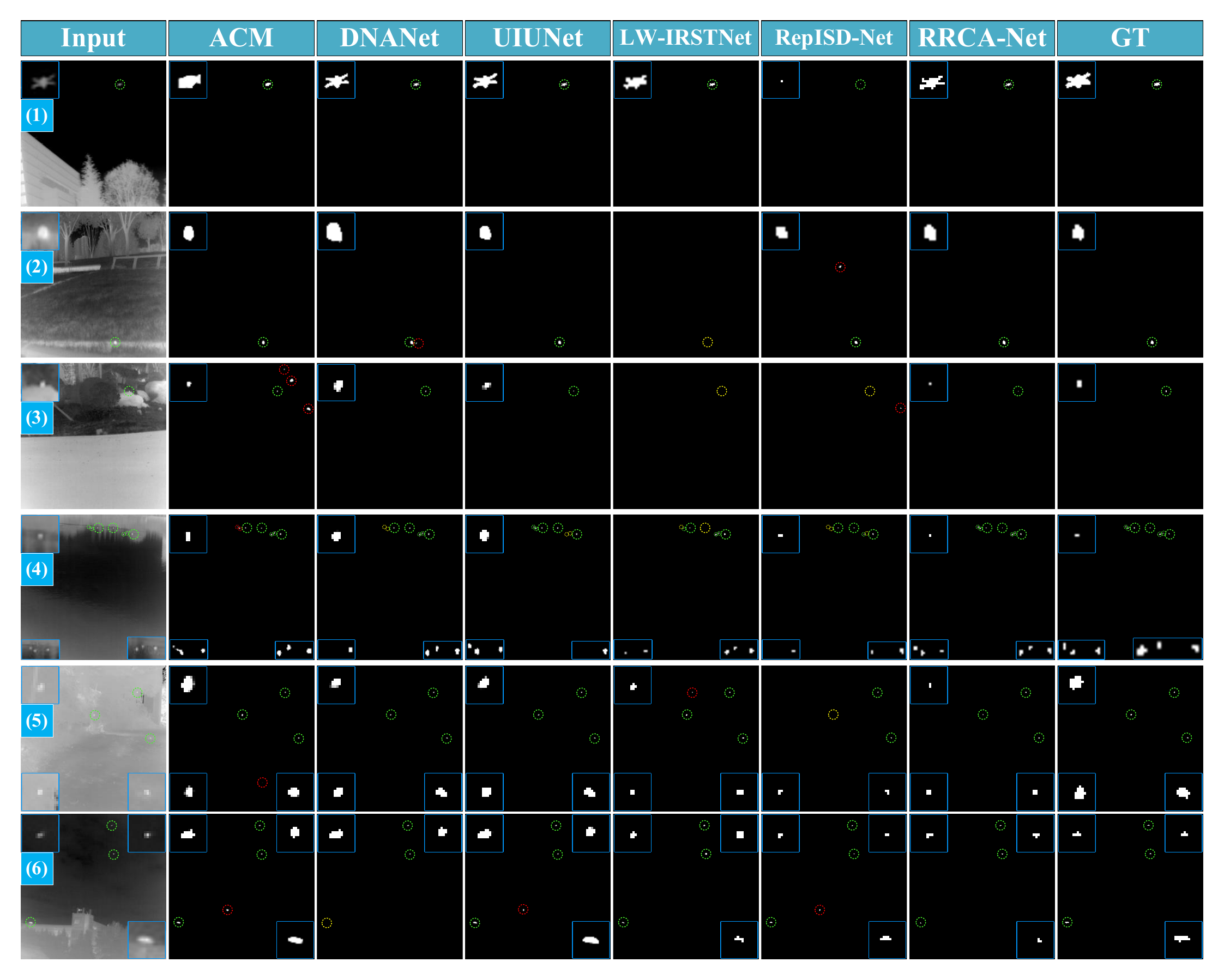}
	\caption{Qualitative results achieved by different SIRST detection methods. For better visualization, the target area is enlarged in the corners of images. The correctly detected target, false alarm, and miss detection areas are highlighted by \textcolor{green}{green}, \textcolor{red}{red} and \textcolor{yellow}{yellow} dotted circles, respectively. Our RRCA-Net can segment changeable targets with precise target localization and shape segmentation under a lower false alarm rate.}\label{visual2D}
	\end{figure*}

	\begin{figure*}
		\centering
		\includegraphics[width=18.2cm]{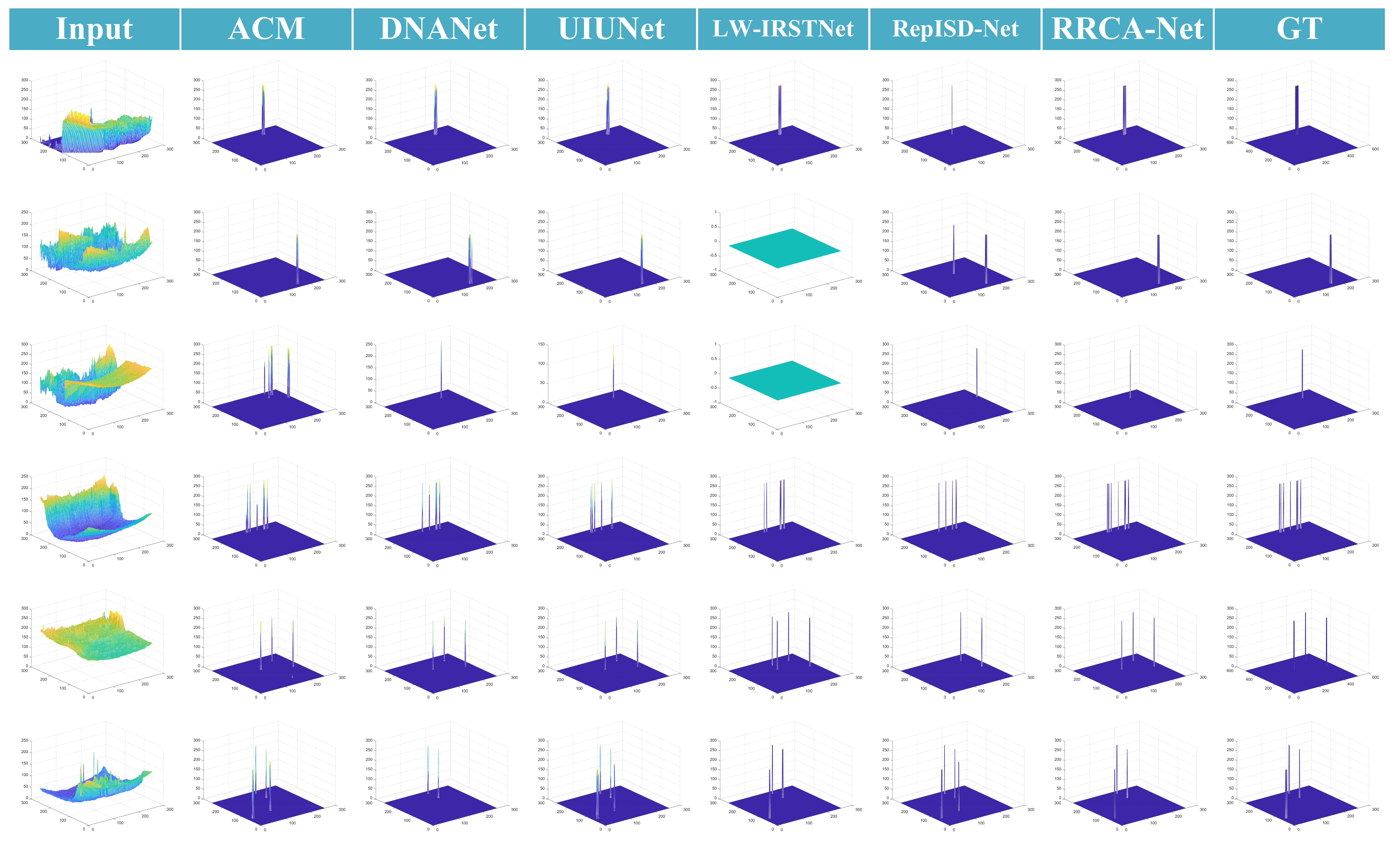}
		\caption{3D visualization results of different methods on different images. }\label{visual3D}
	\end{figure*}
	
	\subsubsection{Qualitative Results}
	Qualitative results of the six representative algorithms are shown in Fig.~\ref{visual2D} and Fig.~\ref{visual3D}. It can be observed that targets have a variety of sizes and types, and they are submerged in low contrast and heavy clutter scenes. Our RRCA-Net can correctly detect the targets and achieve precise contour matching than other CNN-based methods. As illustrated in Fig. 6(1), while these comparison methods can successfully detect the small drone in the air, there are variations in the details description of targets contour. For example, the segmentation results of ACM and RepISD-Net hinder further accurate recognition of target types. In contrast, RRCA-Net finely obtains the drone shape information in images. 
	
	Furthermore, those comparison methods can not suppress the strong noise in heavy clutter background (i.e., image-2 and image-3), LW-IRSTNet has missed the small target in complex background. The situation is similar in the high-light environment (i.e., image-5), low contrast easily causes network misjudgment. In image-6, bright objects that are close to the targets' appearance mislead the network into error detection. For dense targets with varying shapes (i.e., image-4), the comparison methods are more difficult to deal with the fine segmentation of multiple types, and the shape description is not robust. RRCA-Net is effective in improving detection accuracy and reducing miss detection and false alarm. That is because that recurrent iteration can continuously refine contour to match target shape, and correct deviation to make location more accurate. Through the attention mechanism, DIAAM enhances the interactive fusion of low-level and high-level feature maps, avoiding the semantic loss of targets, strengthening the location information and suppressing noise interference. The designed backbone in RRCA-Net can well adapt to various complex scenes, target size and target shape challenges, thus achieves better performance.

	\subsection{Improvements on Existing Methods}
	In Table~\ref{transfer}, we insert our proposed RuCB to three popular IRSTD methods (e.g., ResUNet, AMFU-net and DNANet). After using the module, the detection performance of networks improves (i.e., IoU and $P_{d}$ increase, $F_{a}$ decrease). Especially, the ability to suppress false alarm is greatly improved. That is because that multiple iterations allow the networks to further refine the features and eliminate the interfering information. At the same time, the number of parameters of the network almost remains unchanged. The multiple iterations result in more floating point operations, but the increase in FLOPs is acceptable. Experiments can also verify that our strategy can be transferred to various CNN-based methods and improve detection performance under low computational cost.
	
	\subsection{Ablation Study}\label{SOAT}
	In this subsection, we compare our RRCA-Net with several variants to investigate the potential benefits introduced by our network modules and design choice. We define several symbols for better indication: $C$ represents the all channel numbers of hidden output at different depths; $n$ represents the total recurrence time; $l$ represents the most encoding depth.

	\arrayrulewidth=0.75pt
	\begin{table}[]\scriptsize
		\centering
		\renewcommand\arraystretch{1.1}
		\caption{Results achieved by RRCA-Net's variants that change channel and depth on NUAA-SIRST dataset.} \label{Tab_Channel}
		\begin{tabular}{l c c c c c}
			\hline
			\multicolumn{1}{c}{\multirow{2}{*}{Model Channel$/$Depth}} &  \multicolumn{3}{c}{Performance metrics} & \multirow{2}{*}{$\#$Params}  & \multirow{2}{*}{FLOPs}                 \\ \cline{2-4}
			& $IoU$ & $P_d$ &  $F_a$  &   &               \\ \hline
			C=4,8,16,32;$l$=4       & 73.06  & 97.34 & 24.39  & 0.052 & 0.28\\ 
			C=16,32,64,128;$l$=4    & 74.76  & 95.82 & 21.61  & 0.799 & 4.00      \\ 
			C=8,16,32,64;$l$=4      & \textbf{75.41}  & \textbf{98.10} & \textbf{7.558}  & 0.202 & 1.04     \\ 
			C=8,16,32;$l$=3         & 74.11  & 96.96 & 13.98  & 0.078 & 1.02           \\ 
			C=8,16,32,64,128;$l$=5  & 75.39  & 96.96 & 20.82  & 0.785 & 1.05           \\
			C=4,8,16,32,64;$l$=5    & 72.60  & 98.10 & 23.39  & 0.200 & 0.28   \\ 
			C=16,32,64,128,256;$l$=5  & 75.64  & 96.96 & 22.82  & 3.122 & 4.06          \\ \hline
		\end{tabular}
	\end{table}
	
	\subsubsection{The Backbone}
	As shown in Table~\ref{Tabcomparisonmethod}, we replace the backbone of RRCA-Net and then evaluate the network's performance. The backbone is the residual network with different layers, including ResNet-10, ResNet-18 and ResNet-34. In order to maintain the network depth of four, the number of residual blocks assigned at each depth is (2,1,1,1) for ResNet-10, (3,2,2,2) for ResNet-18 and (4,4,6,3) for ResNet-34. In terms of performance, RRCA-Net with ResNet-18 achieves the balanced performance on three datasets. In terms of number of parameters, RRCA-Net with ResNet-10 has the smallest model size with only 0.103 M parameters. In contrast, the 18-layer backbone introduces $96\%$ more parameters than ResNet-10. However, considering the advantages of RRCA-Net with ResNet-18 in performance and acceptable complexity, we choose it as the main method, and the following experiments are conducted based on this model.

	\arrayrulewidth=0.75pt
	\begin{table}[t]\scriptsize
		\centering
		\captionsetup[table]{labelformat=simple, labelsep=newline, justification=centering, textfont=sc}
		\renewcommand\arraystretch{1.1}
		\caption{Comparison on the number of recurrent iterations in terms of results on the NUAA-SIRST dataset.} \label{Tab_iterations}
		
		\begin{tabular}{c c c c c c}
			\hline
			\multirow{2}{*}{\begin{tabular}[c]{@{}c@{}}Number of iterations \end{tabular}} & \multicolumn{3}{c}{Evaluation Metircs} &   \multirow{2}{*}{$\#$Params}   & \multirow{2}{*}{FLOPs}        \\ \cline{2-4}
			&   IoU &   $P_d$  &  $F_a$  &  &  \\ \hline
			$n=0$   &  73.54  & 96.96   & 16.76   & 0.184  & 0.83   \\ 
			$n=1$   &  74.41  & 96.96   & 12.48   & 0.193  & 0.94   \\ 
			$n=2$   &  \textbf{75.41}  & \textbf{98.10}   & \textbf{7.558}   & 0.202  & 1.04   \\ 
			$n=3$   &  72.62  & 95.82   & 30.66   & 0.211  & 1.15      \\ \hline
		\end{tabular}
	\end{table}
	
	\subsubsection{Channel and Depth}
	To investigate the distribution of parameters and computations across various channels and depths within the network, we perform ablation on relevant elements to find the best performing structure. The Table~\ref{Tab_Channel} shows our experiments results. To ensure that the backbone of all variants is 18 layers, the number of convolution block in each layer is varied, as follows: (3,2,2,2) for four layers, (3,3,3) for three layers, (1,2,2,2,2) for five layers.

	From the comparison of channels, RRCA-Net with C=8,16,32,64 achieves optimal values on three evaluation metrics. The change of the number of channels strongly affects model complexity. When the number of channel is doubled, parameters is increased by 3.9 times and FLOPs is increased by 3.8 times.

	As for depth, even though the five-layer RRCA-Net achieves a higher IoU value than four-layer. But IoU is only 0.23$\%$ more, and network is significantly inferior to the four-layer network in both $P_d$ and $F_a$. Then, RRCA-Net with C=16,32,64,128,256 has too many parameters and operations.

	Then analyzing the distribution of parameters, more parameters are introduced in the deep layer than the shallow layer. That is because, with the deepening of U-shape network structure, channel in each layer gradually doubles. According to formula (\ref{Params}), the number of parameters is strongly related to the number of channels, so it is the key of efficient network to obtain copious feature maps and extract target information by using shallow network structure with few channels. 
    
    As discussed before, C=8,16,32,64 and $l$=4 is the most appropriate channel and depth configurations, respectively.

	\begin{figure*}
	\centering
	\includegraphics[width=18.2cm]{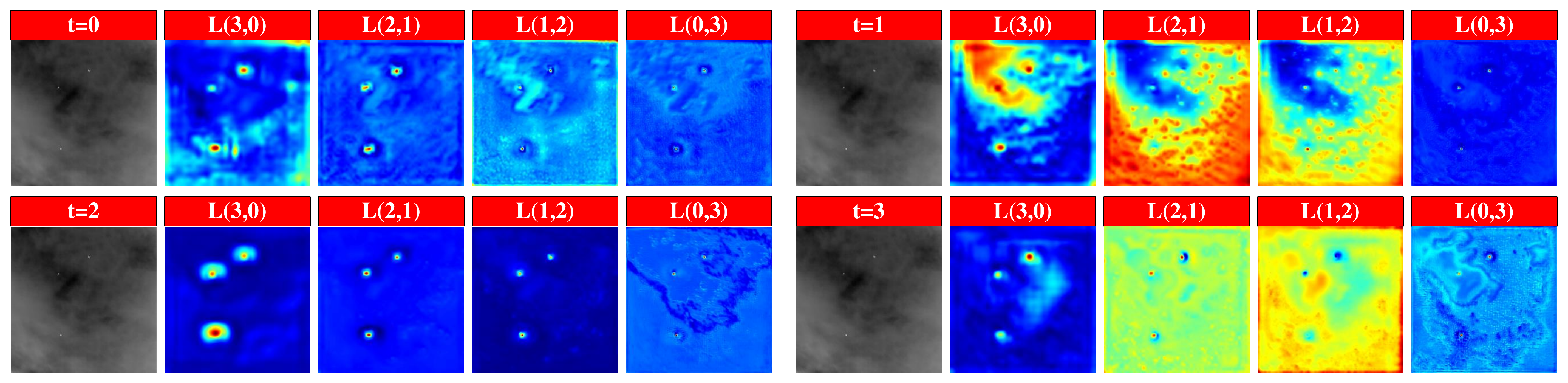}
	\caption{Visualization map of RRCA-Net with different number of iteration. Inappropriate iteration number causes the lose of small target (i.e., \textit{N}=0) and feature overexposure (i.e., \textit{N}=1\&3).}\label{Compare_RuCB}
	\end{figure*}

	\begin{table}[t]\scriptsize
	\centering
	\captionsetup[table]{labelformat=simple, labelsep=newline, justification=centering, textfont=sc}
	\renewcommand\arraystretch{1.1}
	\caption{Results achieved by variants of RRCA-Net and FSM on the NUAA-SIRST dataset. $DS$ means the deep supervision.} \label{Tab_FSM}
	\begin{tabular}{c c c c c c}
		\hline
		\multirow{2}{*}{\begin{tabular}[c]{@{}c@{}}Model \end{tabular}} & \multicolumn{3}{c}{Evaluation Metircs} &   \multirow{2}{*}{$\#$Params}   & \multirow{2}{*}{FLOPs}        \\ \cline{2-4}
		&   IoU &   $P_d$  &  $F_a$  &  &  \\ \hline
		RRCA-Net w/o FSM   &  73.33  & 96.96   & 26.311   & 0.201  & 0.95   \\ 
		RRCA-Net with DS   &  74.25  & 97.72   & 12.763   & 0.201  & 0.95   \\ 
		RRCA-Net   &  \textbf{75.41}  & \textbf{98.10}   & \textbf{7.558}   & 0.202  & 1.04   \\ \hline
	\end{tabular}
	\end{table}

	\begin{figure}
		\centering
		\includegraphics[width=8.8cm]{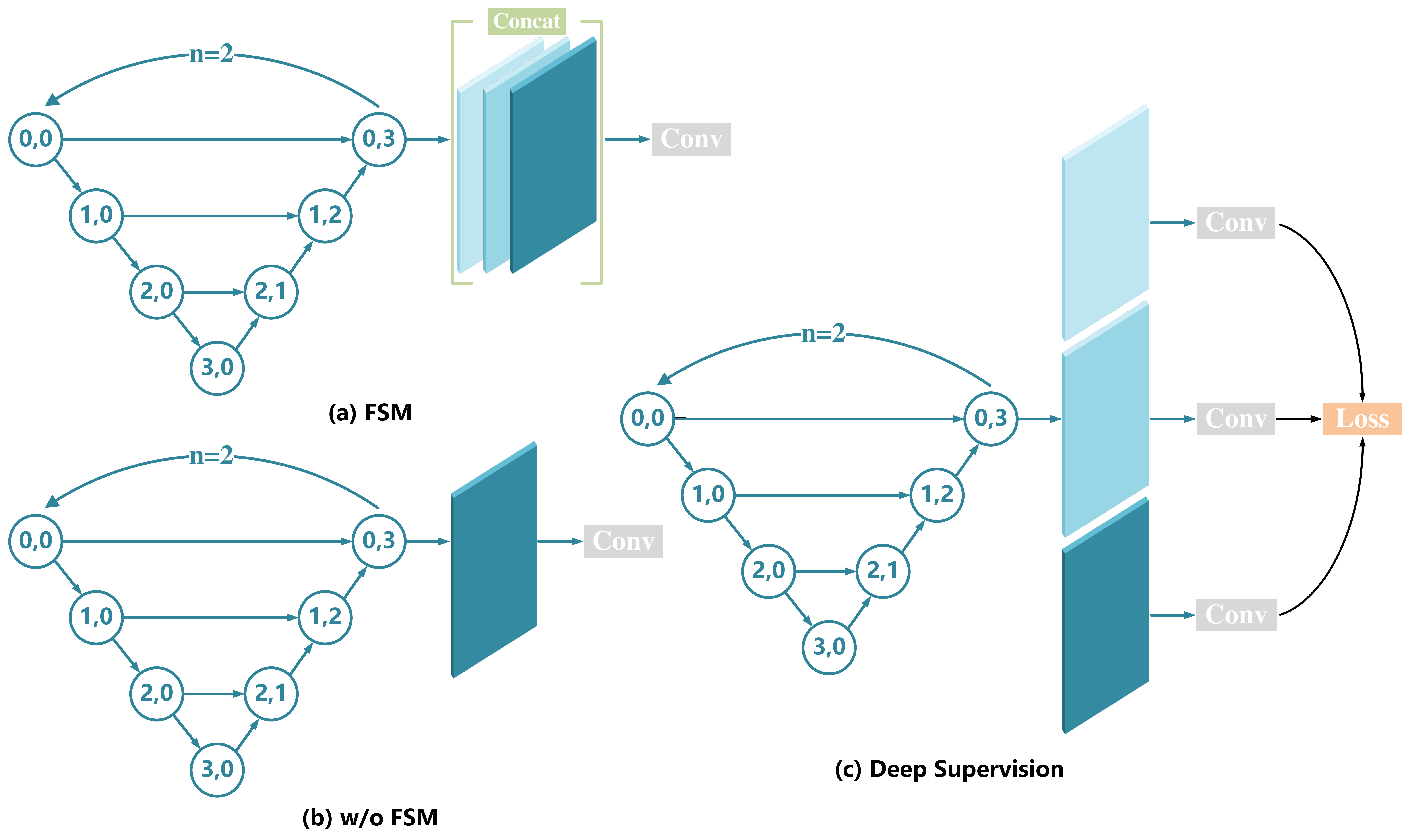}
		\caption{Three variants of FSM. (a) RRCA-Net. (b) RRCA-Net w/o FSM. (c) RRCA-Net with deep supervision, the shade of the color represents feature maps obtained in different recurrent iterations.}\label{Ablation_FSM}
	\end{figure}

	\subsubsection{The Reusable Convolution Block (RuCB)}
	Reusable operation allows the same set of convolution kernels to process feature maps multiple times, achieving progressive depth refinement of detailed features. This is equivalent to functional network deepening, without introducing extra convolution kernel parameters. The reusable convolution block is an important component for achieving recurrent reuse. This block helps to maintain fine-grained features at the finest scale level while eliminate superfluous information. To verify the effectiveness of RuCB, we train and test RRCA-Net with different iterative times $n$.
    
	Table~\ref{Tab_iterations} shows the results achieved by RRCA-Net with changed recurrent times. When $n=0$, network does not reuse convolution kernels, it is the encoder of the plain U-shape network. We can see that segmentation maps is iterated once in the loop, IoU value increases, and false alarm rate decreases. The best performance metrics are obtained at $n=2$, where the $IoU$, $P_d$, and $F_a$ values are improved by $1.87\%$, $1.14\%$ and get decreases of $9.202\times 10^{-6}$ compared to RRCA-Net without RuCB. The reason is that RuCB repeatedly refines features in recurrent iterations to maintain the completeness and comprehensiveness of target information for more advanced performance. However, excessive iteration number (i.e., \textit{N} $\geq$ 3) will make the RRCA-Net suffers from performance degradation, even worse than network without RuCB. We think that too many reusing times will not improve the detection ability, but weaken the network's ability to suppress complex background. 

    Visualization maps in Fig.~\ref{Compare_RuCB} demonstrates the effectiveness of our RuCB. With the appropriate number of recurrent iteration $n=2$, the representation of targets' features is exhaustive, so as to achieve precise location and contour description.

	$\#$Params in Table~\ref{Tab_iterations} have a slight rise with the increase of iterations. That is because that RuCB only reuses the convolution kernels, and new learnable parameters are still added in Batch Normalization layers when iterations occur. And point-wise convolutions in DIAAM also introduce a slight increase in parameters. Each iteration increases $\#$Params by 9K, while FLOPs unsurprisingly increases by 0.11G. Both increase are acceptable.	

	\begin{figure}
		\centering
		\includegraphics[width=8.8cm]{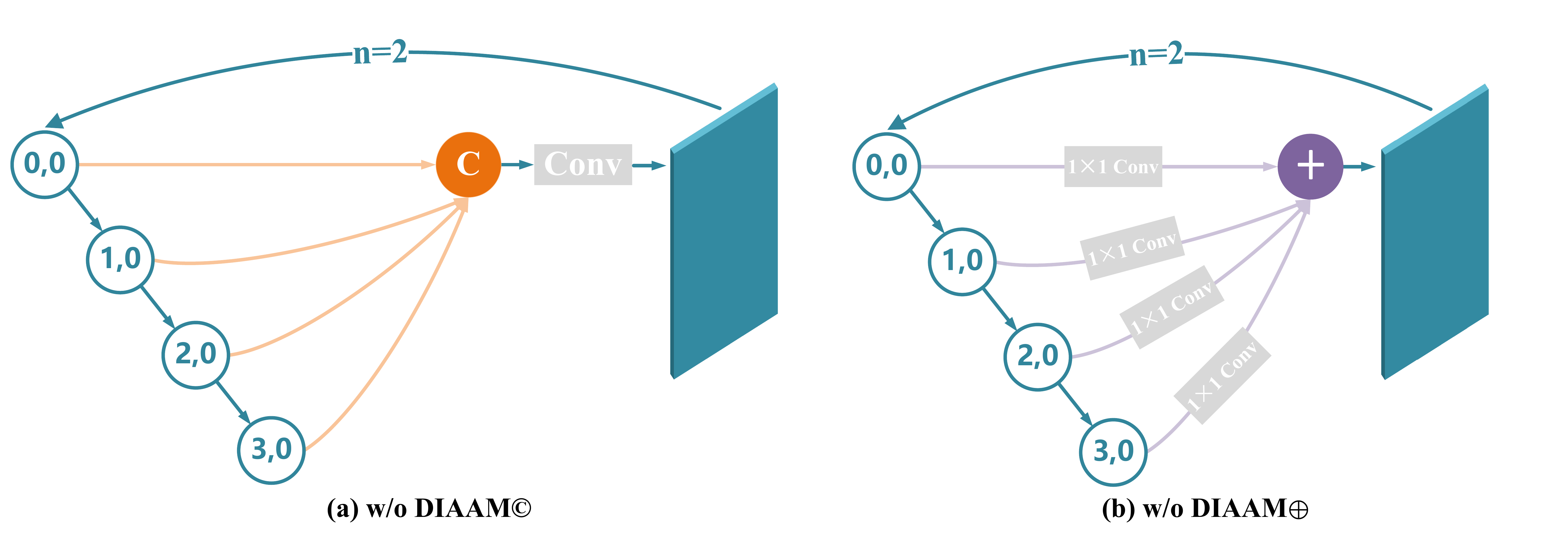}
		\caption{Two variants of DIAAM. (a) RRCA-Net w/o DIAAM$\text{©}$. (b) RRCA-Net w/o DIAAM$\oplus$. \textcolor{orange}{Orange} represents bilinear upsample and concatenate operations, \textcolor{violet}{purple} represents bilinear upsample, $1\times1$ convolution and summation operations.}\label{Ablation_DIAAM}
	\end{figure}
	
	\subsubsection{The Feature Stacking Module (FSM)}
	The feature stacking module is used to stack features, which flows through the encoder-decoder architecture from each iteration and feed them into the last stage block. By fusing coarse-grained and fine-grained features, FSM can enhance the response of small targets. To investigate the benefits of FSM, we compare RRCA-Net with two variants. Results is in Table~\ref{Tab_FSM}.
	
	\begin{itemize}
		\item \textbf{RRCA-Net w/o FSM}: In Fig.~\ref{Ablation_FSM} \textcolor{red}{(b)}, we remove the feature stacking module in this variant and directly process the decoded features from the last iteration to subsequently output prediction.
		\item \textbf{RRCA-Net with Deep Supervision}: We replace FSM with deep supervision in this variant to explore the effectiveness of FSM, as shown in Fig.~\ref{Ablation_FSM} \textcolor{red}{(c)}. Specifically, we process the feature maps obtained in each iteration to generate prediction maps, then calculate the average of each loss values to guide parameter optimization.
	\end{itemize}

	On account of the removal of FSM, the detection performance suffers decreases of 2.08$\%$, 1.14$\%$, and an increase of 18.753 $\times 10^{-6}$ in terms of $IoU$, $P_d$ and $F_a$ on the NUAA-SIRST dataset, respectively. This demonstrates that FSM helps to achieve multi-granularity features fusion. The response from coarse granularity and fine granularity can be extracted in recurrent iterations, then both of them can be fused to generate robust maps as prediction.

	We remove FSM and use deep supervision to process feature maps generated in iterations. Our network suffers decreases of 1.16$\%$, 0.38$\%$ and an increase of 5.205 $\times 10^{-6}$ in terms of $IoU$, $P_d$ and $F_a$. Compared to RRCA-Net w/o FSM, its performance is improved, which verifies the effectiveness of reasonably utilizing many features from multiple iterations. Fusion is more conductive to performance improvement.

	\subsubsection{The Dual Interactive Attention Aggregation Module (DIAAM)}
	
	To reduce model complexity, it is a main factor to minimize the number of convolutions introduced. The dual interactive attention aggregation module is used to replace the decoder structure in the traditional U-shape network. DIAAM realizes the aggregation of shallow features with abundant spatial information and deep features with abundant semantic information in low resource consumption. To study the efficacy of DIAAM, we compare RRCA-Net with several variants that adopt other aggregation methods. In order to ensure fairness, the encoders are consistent for all models and iterations $n=2$. Fig.~\ref{Ablation_DIAAM} shows these variants.

	\begin{itemize}
		\item \textbf{RRCA-Net w/o DIAAM (Concatenate)}: We remove the dual interactive attention aggregation  module in this variant and upsample multi-layer features extracted in encoder to the same resolution. Then, these same resolution maps are concatenated along channel dimension. Subsequently, it is fused by a residual convolution block.
		\item \textbf{RRCA-Net w/o DIAAM (Element-wise summation)}: After removing the dual interactive attention aggregation module, we upsample the features of each layer in encoder and process them with $1\times 1$ convolution to unify channel number. Subsequently, multi-layer features are fused by the element-wise summation. 
		\item \textbf{RRCA-Net w/o CA}:  We remove the channel attention operation in high-level branch of DIAAM to analyze its contribution.
		\item \textbf{RRCA-Net with CA4}: We change the scaling ratio of the shared MLP in channel attention. The ratio factor is set to 4. The original factor of RRCA-Net is 8.
		\item \textbf{RRCA-Net with re-weight}: We change the fusion strategy between two branches. Specifically, generating wights from high-level features and then weighting the shallow-level features precessed by channel attention.
	\end{itemize}

		\begin{table}[t]\scriptsize
		\centering
		\renewcommand\arraystretch{1.1}
		\caption{Results achieved by main variants of RRCA-Net and DIAAM on the NUAA-SIRST datasets. $\text{©}$ and $\oplus$ respectively mean concatenate and element-wise summation as feature fusion methods. $CA$ means the channel attention mechanism.} \label{Tab_DIAAM}
		\begin{tabular}{c c c c c c}
			\hline
			\multirow{2}{*}{\begin{tabular}[c]{@{}c@{}}Model \end{tabular}} & \multicolumn{3}{c}{Evaluation Metircs} &   \multirow{2}{*}{$\#$Params}   & \multirow{2}{*}{FLOPs}        \\ \cline{2-4}
			&   IoU &   $P_d$  &  $F_a$  &  &  \\ \hline
			RRCA-Net w/o DIAAM$\text{©}$    &  74.11  & 96.58   & 25.669   & 0.199  & 1.60   \\ 
			RRCA-Net w/o DIAAM$\oplus$      &  73.26  & 96.58   & 26.311   & 0.190  & 0.98   \\ 
			RRCA-Net w/o CA                 &  73.17  & 96.20   & 20.892   & 0.201  & 1.04   \\ 
			RRCA-Net with CA4 				&  75.17  & 96.20   & 18.967   &0.202   &1.04		\\
			RRCA-Net with re-weight			&73.29 &96.96	&24.100	&0.202		&1.04		\\
			RRCA-Net   &  \textbf{75.41}  & \textbf{98.10}   & \textbf{7.558}   & 0.202  & 1.04   \\ \hline
		\end{tabular}
	\end{table}

	As shown in Table~\ref{Tab_DIAAM}, RRCA-Net w/o DIAAM$\oplus$ suffers decreases of 2.15\%, 1.52\%, and a increase of 18.753 $\times 10^{-6}$ in terms of $IoU$, ${P}_{d}$, and ${F}_{a}$ on the NUAA-SIRST dataset. Similar results can also be observed for RRCA-Net w/o DIAAM$\text{©}$. The reason is that DIAAM gradually aggregates the upper-layer and lower-layer features to maintain the elaborate information such as centroid location and shape contour. Conversely, upsampling is prone to generate deviations.

	RRCA-Net w/o CA suffer decreases of 2.24$\%$, 1.9$\%$ and an increase of 13.334 $\times 10^{-6}$ in terms of $IoU$, ${P}_{d}$, and ${F}_{a}$ on the NUAA-SIRST dataset. That is because that channel attention mechanism in DIAAM can well exploit significant channels to enhance the response of features. RRCA-Net with CA4 also shows a decrease in three evaluation metrics. Appropriate scaling facilitates the preservation of feature information. RRCA-Net with re-weight suffers decreases of 2.12$\%$, 1.14$\%$ and an increase of 16.542 $\times 10^{-6}$ in terms of $IoU$, ${P}_{d}$, and ${F}_{a}$. This suggests that using shallow-level features rich in spatial information to weight high-level features improves the network's ability to locate targets.  As for model complexity, these models differ sightly in $\#$Params. While the concatenate operation brings 0.56 G FLOPs, a large number of channels cause the increase.
	
	\begin{figure}[t]
		\centering
		\includegraphics[width=8.8cm]{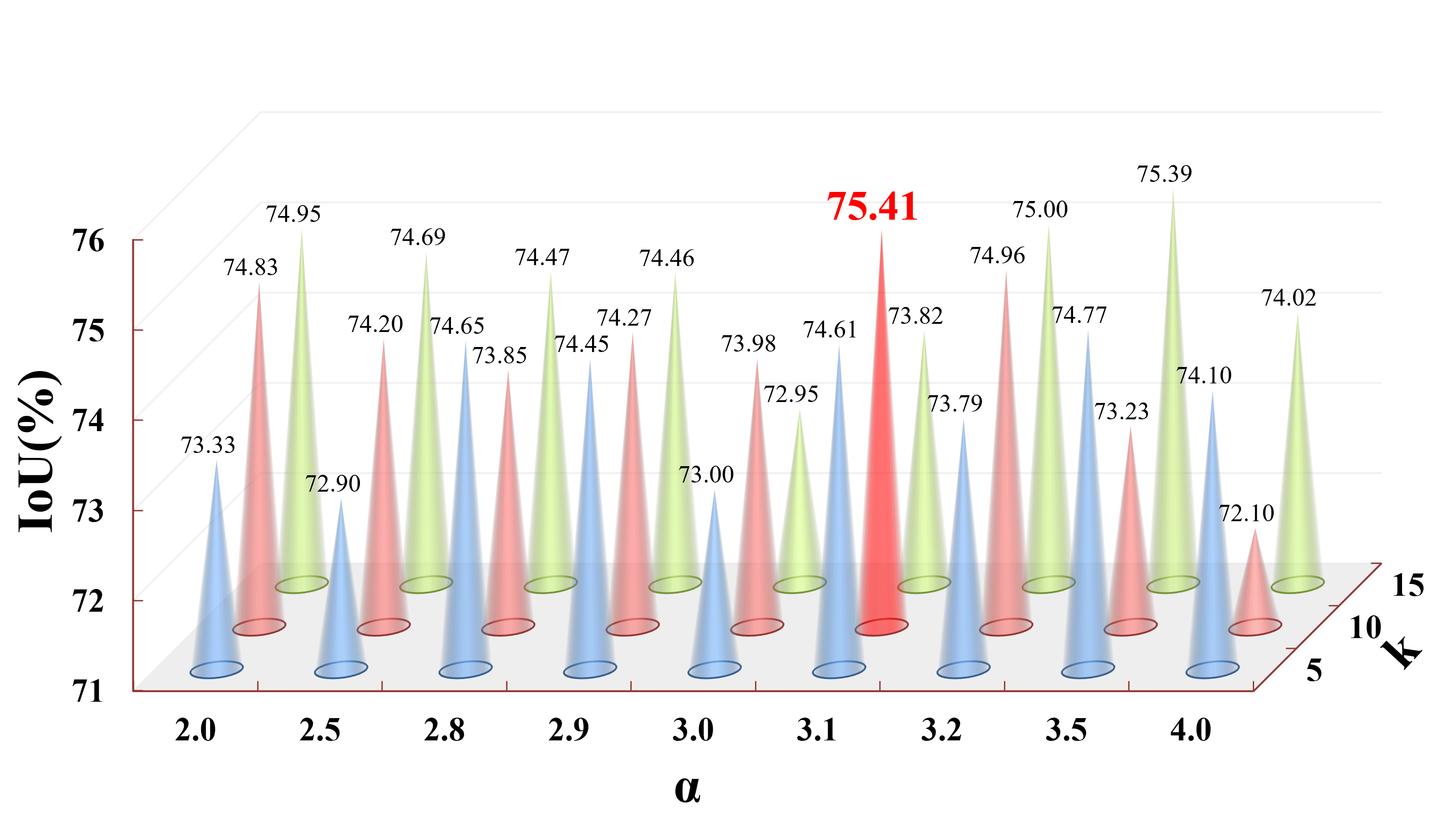}
		\caption{IoU values achieved by RRCA-Net using DpT-k loss with varying hyperparameters.}\label{lossParams}
	\end{figure}
	
	\arrayrulewidth=0.75pt
	\begin{table}[t]\scriptsize
		\centering
		\captionsetup[table]{labelformat=simple, labelsep=newline, justification=centering, textfont=sc}
		\renewcommand\arraystretch{1.1}
		\caption{Results achieved by RRCA-Net on the NUAA-SIRST dataset, which uses DpT-k loss with changeable hyperparameters.} \label{Tab_DptkLoss}
		
		\begin{tabular}{c c c c c c c}
			\hline
			\multirow{2}{*}{\begin{tabular}[c]{@{}c@{}}Hyperparameters \end{tabular}} & \multicolumn{2}{c}{$k=5$} &   \multicolumn{2}{c}{$k=10$}   & \multicolumn{2}{c}{$k=15$}        \\ \cline{2-7}
			&   $P_d$  &  $F_a$  & $P_d$ & $F_a$ & $P_d$ & $F_a$ \\ \hline
			$\alpha=2.0$   & 95.44   & 17.041   & 95.06   & 18.040  & 96.96  & 12.621 \\ 
			$\alpha=2.5$   & 96.20   & 20.606   & 96.96   & 21.320  & 96.96  & 11.337   \\ 
			$\alpha=2.8$   & 97.34   & 19.680   & 96.20   & 14.118  & 97.72  & 19.537   \\ 
			$\alpha=2.9$   & 96.58   & 10.909   & 96.96   & 9.626   & 96.96  & 12.264   \\  
			$\alpha=3.0$   & 96.58   & 11.979   & 96.20   & 13.191  & 94.68  & 23.530   \\ 
			$\alpha=3.1$   & 97.34   & 20.606   & \textbf{98.10}   & \textbf{7.558} & 96.96  & 14.688   \\ 
			$\alpha=3.2$   & 95.82   & 18.610   & 97.34   & 14.974  & 97.34  & 12.193   \\ 
			$\alpha=3.5$   & 96.96   & 20.535   & 96.20   & 16.614  & 95.82  & 17.113   \\ 
			$\alpha=4.0$   & 96.96   & 14.688   & 96.20   & 21.676  & 97.34  & 11.052   \\ \hline
		\end{tabular}
	\end{table}

	\subsection{Benefits of DpT-k Loss}\label{SOAT}	
	In this section, we evaluate the benefits of our designed loss function for IRSTD. Specifically, we investigate the effect of hyperparameters on the ability of DpT-k loss, and find the best pair. To analyze the internal essence why DpT-k loss works, we disassemble and combine DpT-k loss function to form several variants. Then, we respectively train RRCA-Net with these variants and several typical loss functions to compare the convergence ability.
	
	First, we change two hyper-parameters, $\alpha$ and $k$, which mainly affect the ability of DpT-k loss, to find the optimal loss function. From Fig.~\ref{lossParams} and Table~\ref{Tab_DptkLoss}, when $\alpha=3.1$ and $k=10$, DpT-k loss supports RRCA-Net to achieve significant detection performance improvement through training.

\begin{figure}
		\centering
		\includegraphics[width=8.8cm]{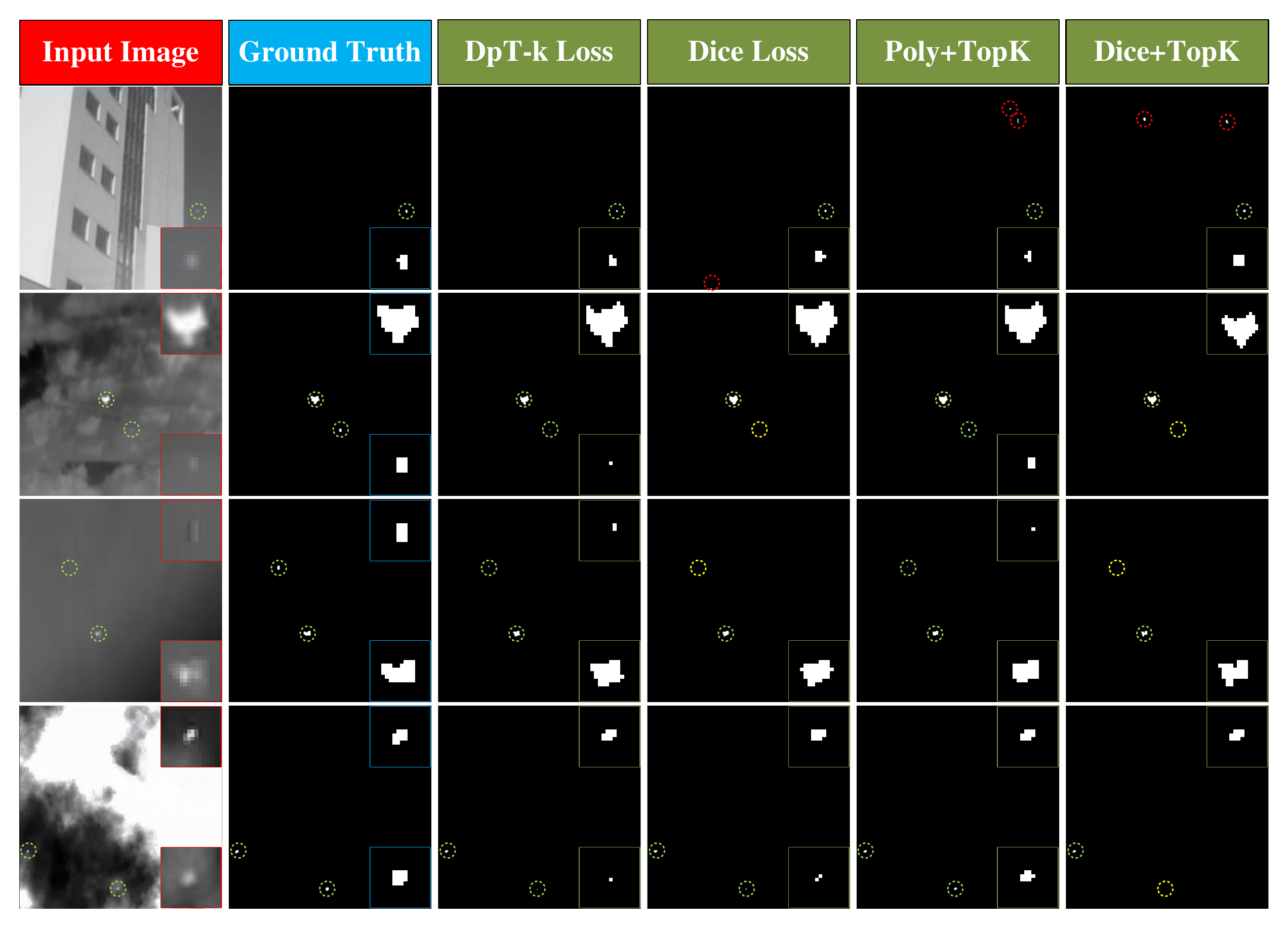}
		\caption{Qualitative results achieved by RRCA-Net which is trained with different loss functions.}\label{Visual_loss}
	\end{figure}

	In Table~\ref{Tab_loss}, we can find that the single losses (e.g., Dice Loss\cite{GDice}, SoftIoU Loss\cite{softiou}) perform worse than the joint losses on three metrics. As to joint losses, Dice$+$Focal loss \cite{DiceFocal} and FocalIoU loss \cite{32-mtu-Net} show the conventional performance and does not reach the optimum. Dice$+$Top-k loss also achieves the highest IoU value close to DpT-k loss. But $P_d$ and $F_a$ achieved by Dice$+$Top-k loss are worse.  DpT-k loss achieves the advanced detection performance, especially obtaining a low false alarm. As shown in Fig.~\ref{Visual_loss}, RRCA-Net trained with DpT-k loss has a strong ability to locate dim targets and suppress the false alarms. That is because that DpT-k loss can combine the advantages of Dice loss, Poly loss and Top-k loss, the approach allows loss function to help RRCA-Net focuses on hard samples and multi-scale targets. Meanwhile the combination of shape constraints at physical (i.e., the contribution of Dice) and distribution constraints at the mathematical level (i.e. the contribution of Poly and Top-k) guide the network to converge more stably and get better performance.

	\begin{table}[]\scriptsize
		\centering
		\renewcommand\arraystretch{1.1}
		\caption{Results achieved by  RRCA-Net with different loss functions on the NUAA-SIRST datasets.} \label{Tab_loss}
		\begin{tabular}{c c c c}
			\hline
			\multirow{2}{*}{\begin{tabular}[c]{@{}c@{}}Loss function \end{tabular}} & \multicolumn{3}{c}{Evaluation Metircs}         \\ \cline{2-4}
			&   IoU &   $P_d$  &  $F_a$   \\ \hline
			SoftIoU Loss    & 73.63   & 96.20   &  14.902    \\
			Dice Loss       & 73.41   & 96.58   &  15.045    \\ 
			Dice Loss + Focal Loss & 74.13 & 96.58& 21.034 \\
			FocalIoU Loss & 72.28 & 95.82 & 11.765 \\
			Top-k Loss ($k=10$)      & 73.85   & 97.72   &  16.328  \\ 
			Poly Loss ($\alpha=3.1$)       & 72.50   & 95.82   &  11.052    \\ 
			Poly Loss ($\alpha=3.1$) + Top-k Loss ($k=10$)  & 72.95   & 97.34   &  22.817    \\ 
			Dice Loss + Top-k Loss ($k=10$)  & 75.40   & 96.96   &  15.401    \\ 
			DpT-k Loss ($\alpha=3.1, k=10$)  &  \textbf{75.41}  & \textbf{98.10}   & \textbf{7.558}   \\ \hline
		\end{tabular}
	\end{table}

	\section{CONCLUSION}\label{SecConclusion}
	In this paper, we  propose a RRCA-Net to achieve high-performance IRST detection with low computation cost. Different from previous lightweight methods, we are inspired by the reusing mechanism and designed a reusable network with cascaded attention module to achieve high-level feature refinement and enhancement, respectively. Moreover, we develop a target characteristic-inspired loss function to guide the convergence toward precise location and contour description. Experiments on three datasets have shown that RRCA-Net achieves comparable performance to the state-of-the-art methods, and introduces consistent performance improvement for several popular IRSTD methods. However, an obvious limitation of recurrent reuse is that FLOPs will increase. In the future, we will consider reusing the encoder as a decoder to further reduce the model size. And we attempt to introduce reuse operations into the transformer structure.

	\bibliographystyle{IEEEtran}
	\bibliography{DNANet}
	\begin{IEEEbiography}[{\includegraphics[width=1in,height=1.25in,clip,keepaspectratio]{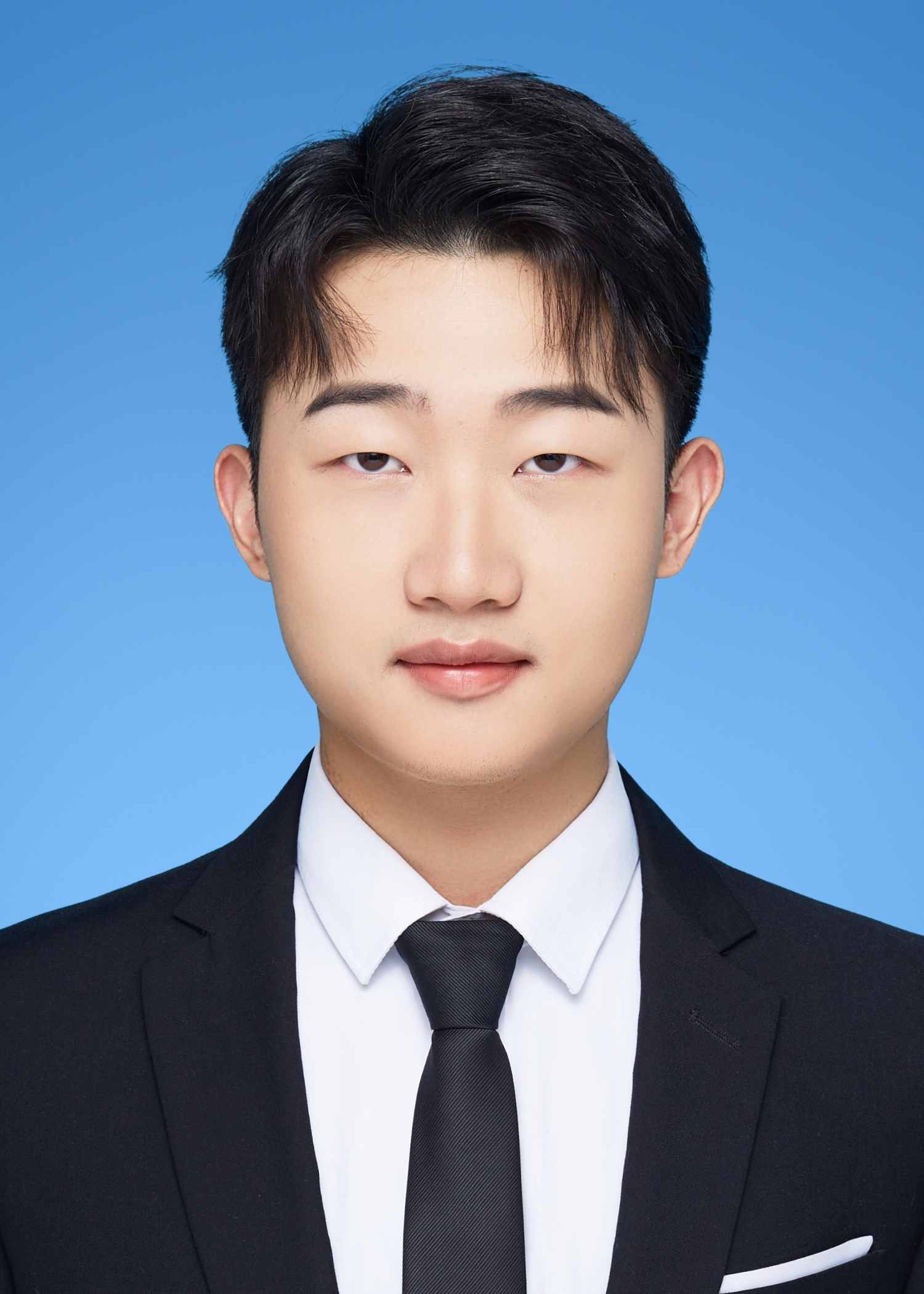}}]{Yongxian Liu} received the B.E. degree in Electromagnetic Field and Wireless Technology from the Xidian University, Xi’an, China, in 2023. He is currently working toward the M.E. degree in information and communication engineering with the National University of Defense Technology(NUDT), Changsha, China. His research interests focus on infrared small target detection and tracking.
	\end{IEEEbiography}
	\begin{IEEEbiography}[{\includegraphics[width=1in,height=1.25in,clip,keepaspectratio]{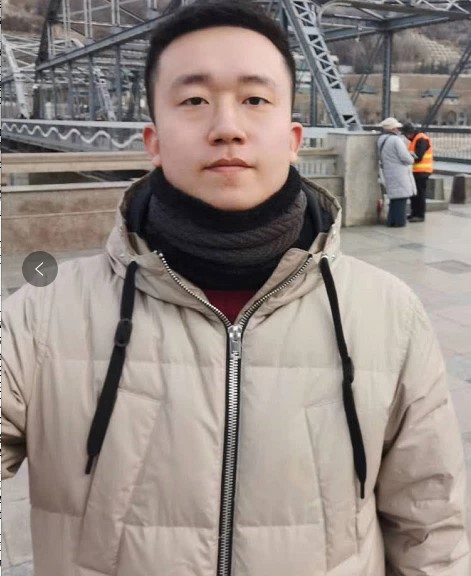}}]{Boyang Li} received the Ph.D and Master degrees from NUDT and National Defense Technology Innovation Institute in 2024 and 2020, respectively. Before that, He received the B.E. degree from Tianjin University in 2017. Currently, he is an assistant professor with the College of Electronic Science and Technology, NUDT. His research interests focus on optical image processing, interpretation and application, particularly on infrared small target detection, weakly supervised semantic segmentation, and neural network compression and accelerating.
	\end{IEEEbiography}
	\begin{IEEEbiography}[{\includegraphics[width=1in,height=1.25in,clip,keepaspectratio]{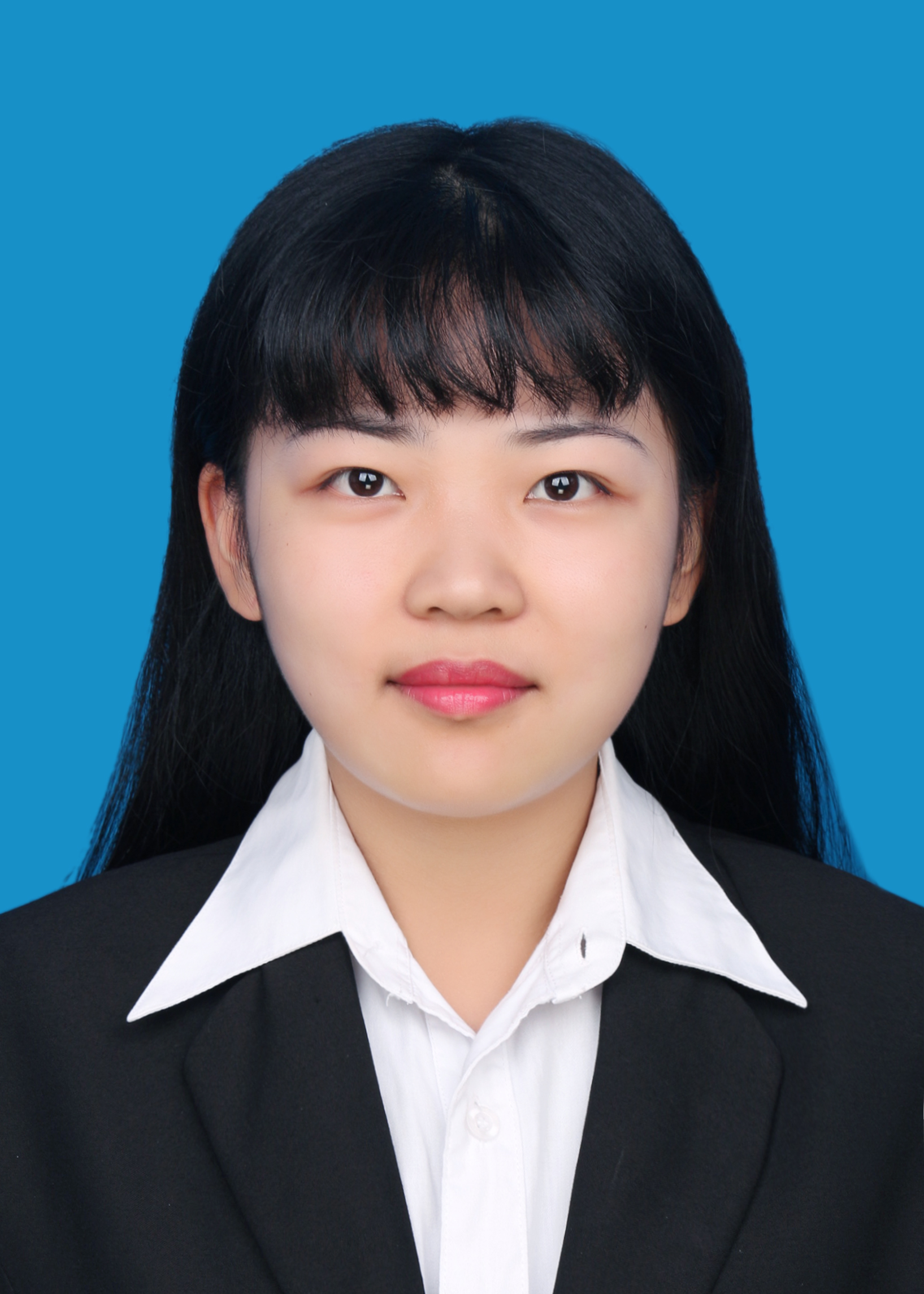}}] {Ting Liu} received the B.E. degree in electrical engineering and automation from Hunan Institute of Engineering , Xiangtan, China, in 2017, the M.E. degree in control engineering from Xiangtan University (XTU), Xiangtan, China, in 2020, and the Ph.D. degree in information and communication engineering from the National University of Defense Technology (NUDT), Changsha, China, in 2024. She is currently a lecturer with the College of Automation and Electronic Information, Xiangtan University. Her research interests focus on signal processing and image processing, particularly on infrared small target detection and satellite video object detection.
	\end{IEEEbiography}
	\begin{IEEEbiography}[{\includegraphics[width=1in,height=1.25in,clip,keepaspectratio]{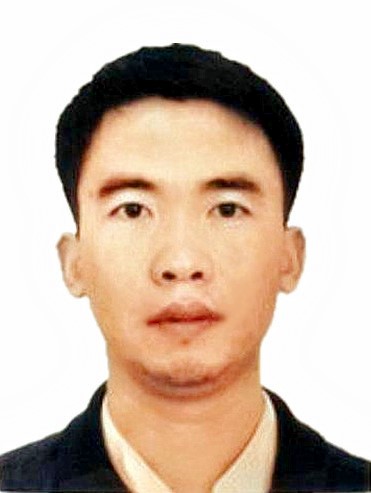}}]{Zaiping Lin} received the B.Eng. and Ph.D. degrees from the National University of Defense Technology (NUDT) in 2007 and 2012, respectively. He is currently an Associate Professor with the College of Electronic Science and Technology, NUDT. His current research interests include infrared image processing and signal processing.
	\end{IEEEbiography}
	
	\begin{IEEEbiography}[{\includegraphics[width=1in,height=1.25in,clip,keepaspectratio]{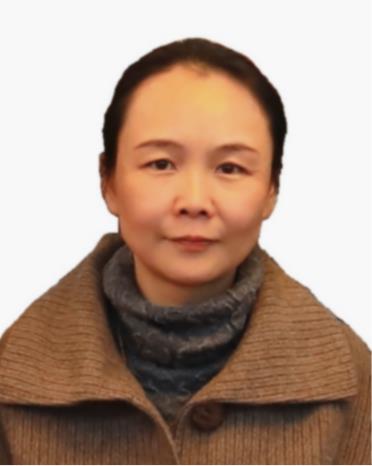}}]{Wei An} received the Ph.D. degree from the National University of Defense Technology (NUDT), Changsha, China, in 1999. She was a Senior Visiting Scholar with the University of Southampton, Southampton, U.K., in 2016. She is currently a Professor with the College of Electronic Science and Technology, NUDT. She has authored or co-authored over 100 journal and conference publications. Her current research interests include signal processing and image processing.
	\end{IEEEbiography}
\end{document}